\documentclass[letterpaper,twocolumn,anonymous,10pt]{article}
\usepackage{usenix2019_v3}

\usepackage{tikz}
\usepackage{amsmath}

\usepackage{filecontents}
\usepackage{xspace}
\usepackage{threeparttable}
\usepackage{makecell}
\usepackage{url}
\usepackage{cleveref}
\usepackage{enumitem}

\usepackage{tikz}
\usepackage{threeparttable}
\usepackage{soul}
\usepackage{colortbl}
\usepackage{multirow}
\usepackage{tabularx}
\usepackage{pifont, mathtools}
\usepackage[overload]{empheq}
\usepackage{nccmath}
\usepackage{enumitem}
\usepackage{cases}
\usepackage{bbding}
\usepackage{graphicx}
\usepackage{pifont}
\usepackage{wasysym}
\usepackage{mathtools}
\usepackage{float}
\usepackage{subcaption}
\usepackage{adjustbox}
\usepackage{makecell}%
\usepackage{tikz}
\usepackage{booktabs}
\usepackage{cleveref}
\usepackage{xspace}
\usepackage{comment}
\usepackage[normalem]{ulem}
\usepackage{enumitem}
\usepackage{listings}
\usepackage{color}
\usepackage{caption}
\usepackage{textcomp}
\usepackage{balance}
\usepackage{hyperref}
\usepackage{booktabs}
\usepackage{tabularx}
\usepackage{caption}
\usepackage{tikz}

\usepackage{pifont}
\newcommand{\xmark}{\ding{55}}

\usepackage{enumitem} 
\setlist{noitemsep,topsep=2pt,parsep=2pt,partopsep=0pt}
\usepackage{cleveref}
\crefname{section}{}{\S\S}
\crefdefaultlabelformat{\S#2#1#3}

\definecolor{red}{rgb}{1,0,0}
\definecolor{gray}{rgb}{0.5,0.5,0.5}

\newcommand{\fyc}[1]{{\color{black} #1}}

\newcommand\hmm[1]{\ifnum\ifhmode\spacefactor\else2000\fi>1000 \uppercase{#1}\else#1\fi}

\newcommand{\op}{operation\xspace}
\newcommand{\ops}{operations\xspace}

\newcommand{\sys}{Echo\xspace}
\newcommand{\bs}{Proteus\xspace}
\newcommand{\ff}{FlexFlow\xspace}

\newcommand{\wg}{\hmm{e}xecution graph\xspace} %
\newcommand{\Ep}{Ex-situ simulation\xspace} %
\newcommand{\ep}{ex-situ simulation\xspace} %

\newcommand{\extractor}{\hmm{w}orkload tracer\xspace} %
\newcommand{\cm}{CC estimator\xspace} %
\newcommand{\ma}{\hmm{t}imeline composer\xspace} %
\newcommand{\va}{\hmm{v}alidator\xspace} %
\newcommand{\nccl}{NCCL\xspace}
\newcommand{\ar}{\texttt{all-reduce}\xspace} %
\newcommand{\ag}{\texttt{all-gather}\xspace} %
\newcommand{\rs}{\texttt{reduce-scatter}\xspace} %
\newcommand{\send}{\texttt{send}\xspace} %
\newcommand{\recv}{\texttt{receive}\xspace}  %

\usepackage[compact]{titlesec}

\usepackage{amsmath}
\usepackage{amssymb}

\begin{document}

\title{\Large \bf \sys: Simulating Distributed Training At Scale}

\author{
{\rm Yicheng Feng$^{\text{1}}$}\enskip
{\rm Yuetao Chen$^{\text{1}}$}\enskip
{\rm Kaiwen Chen$^{\text{1}}$}\enskip
{\rm Jingzong Li$^{\text{2}}$}\enskip
{\rm Tianyuan Wu$^{\text{3}}$}\enskip
\\
{\rm Peng Cheng$^{\text{4}}$}\enskip
{\rm Chuan Wu$^{\text{5}}$}\enskip
{\rm Wei Wang$^{\text{3}}$}\enskip
{\rm Tsung-Yi Ho$^{\text{1}}$}\enskip
{\rm Hong Xu$^{\text{1}}$}\enskip
\\
\\
$^{\text{1}}$The Chinese University of Hong Kong\enskip
$^{\text{2}}$The Hang Seng University of Hong Kong\enskip
$^{\text{3}}$HKUST\enskip\\
$^{\text{4}}$Microsoft Research \enskip
$^{\text{5}}$The University of Hong Kong
}

\maketitle

\begin{abstract}

Simulation offers unique values for both enumeration and extrapolation purposes, and is becoming increasingly important for managing the massive machine learning (ML) clusters and large-scale distributed training jobs.
In this paper, we build \sys to tackle three key challenges in large-scale training simulation: 
(1) tracing the runtime training workloads at each device in an ex-situ fashion so we can use a single device to obtain the actual execution graphs of 1K-GPU training, (2) accurately estimating the collective communication without high overheads of discrete-event based network simulation, and (3) accounting for the interference-induced computation slowdown from overlapping communication and computation kernels on the same device.
\sys delivers on average 8\% error in training step---$\sim$3x lower than state-of-the-art simulators---for GPT-175B on a 96-GPU H800 cluster with 3D parallelism on Megatron-LM under 2 minutes.

\end{abstract}

\section{Introduction} \label{sec:introduction}

The unprecedented success of large language models (LLMs) has been driven in large part by the large-scale infrastructures that allow the model and training dataset to scale. 
Organizations have constructed massive clusters with tens of thousands of GPUs to train models with hundreds of billions or even trillions of parameters~\cite{jiang2024megascale,goyal2017accurate,llama3-meta}, in a continuous stride to improve model capabilities as promised by the scaling law~\cite{kaplan2020scaling}.

Training LLMs on such massive scales consumes substantial time and financial resources. On the other hand distributed training is inherently complex and rapidly evolving, involving a myriad of algorithm/model innovations (linear attention \cite{katharopoulos2020transformers}), parallelism strategies~\cite{li2020pytorch,zhao2023pytorch,shoeybi2019megatron,ren2021zero}, operation and kernel optimization (kernel fusion~\cite{dao2022flashattention}), and hardware designs \cite{memryx}. 
As a result, accurate simulation of distributed training has started to garner attention as an essential tool to effectively manage training infrastructures and systems~\cite{SimAI2025NSDI,duan2024proteus,won2023astra,lu2023distsim,hu2022dpro}. 

Broadly, simulation is useful for two main purposes: enumeration and extrapolation.
In distributed training context, simulation can be used to enumerate combinations of parallelism strategies, optimization techniques, and hardware configurations that are available in the current cluster, to derive the optimal training plan for a given job and to determine efficient resource allocation schedules across jobs.
Simulation is also useful (perhaps more so) to extrapolate beyond what is currently available, which is paramount for strategic decision making such as capacity planning~\cite{jiang2024megascale,shoeybi2019megatron} that involve many what-if questions with significant impact. 
For example, what speed-up can be achieved by scaling the current cluster by a factor of 3, or by increasing the network bandwidth by 2x?
This also greatly facilitates the development of new optimizations, which only need to be prototyped on a small scale for the simulator to extrapolate its potential benefits on a large scale quantitatively.

Prior work has made salient progress in training simulation~\cite{SimAI2025NSDI,duan2024proteus,won2023astra,lu2023distsim,hu2022dpro}. 
Yet, they fall short in three key aspects that hinder their capabilities in supporting both enumeration and extrapolation practically.

First, modern large-scale training often employs 3D parallelism, a combination of data parallelism (DP), tensor parallelism (TP), and pipeline parallelism (PP), to break down the large models along different dimensions into sub-models that fit into a device's memory and maximize cluster resource utilization. 
To accurately simulate 3D parallelism, one needs to obtain the training workloads, which encompass the device- or rank-specific execution graphs detailing the computation and communication \ops and their dependencies.
Existing work \cite{hu2022dpro,lu2023distsim,luo2022srifty,zhu2020daydream,SimAI2025NSDI} requires a full-scale cluster deployment of the job to trace the training workloads from each and every device at runtime.
This is because each device builds its own execution graph independently in parallel during job initialization to most efficiently carry out training of its unique sub-model according to the 3D parallelism setting and various optimization techniques (e.g. kernel fusion).
The \textit{in-situ} tracing approach is clearly very costly and sometimes even impractical for extrapolation usecases.

{Second}, simulating collective communication (CC) (e.g. \ar) is critical because (1) it pieces all the devices and nodes together according to the various parallelisms, and (2) its performance can often be the bottleneck \cite{duan2024efficient, qi2023zero, jiang2024megascale, huang2019gpipe}. 
However, most existing work~\cite{duan2024proteus, won2023astra,hu2022dpro,lu2023distsim,zhu2020daydream} rely on a coarse-grained $\alpha$-$\beta$ model \cite{valiant1990bridging} without considering the actual communication patterns and optimizations in the implementation, resulting in significant simulation errors.  
On the other hand, more recent work such as SimAI \cite{SimAI2025NSDI} explicitly simulates each peer-to-peer send and receive primitive of a CC kernel using a packet-level event-based network simulator such as ns-3 \cite{riley2010ns}.
This fine-grained approach provides high accuracy at the cost of prohibitive overheads even at a medium scale: simulating a 128-GPU job takes SimAI over 2 hours with an optimized ns-3 implementation (\cref{subsec:challenges}). 

{Third}, we observe that overlapping communication and computation \ops, an optimization widely used to improve efficiency \cite{jiang2024megascale,chen2024centauri,zhang2017poseidon}, incurs non-negligible slowdowns to the computation due to contention for shared resources (e.g., cache, DRAM bandwidth) \cite{hwang2023ark,duan2024proteus}.
This, however, has largely been overlook in prior work thus far.
Simulating interference-induced slowdown is particularly challenging since it depends on many idiosyncrasies ranging from scheduling logic of the ML framework and the underlying library to hardware-specific implementation optimizations.
Much of these details are vendor-proprietary and inaccessible.

In this work, we propose \sys, a practical simulation platform for large-scale distributed training. 
\sys is built with three key design choices.
(1) It adopts an \textit{ex-situ} tracing approach to accurately capture training workloads using just a single device, avoiding the need for a full-scale cluster deployment. The key technical idea here is to transform the parallel initialization process at each rank into a sequential one, allowing a single device to act as each rank iteratively to trace the corresponding execution graph.
(2) It employs white-box modeling to simulate a CC kernel's performance with four parts: connection setup overhead, intra-server and inter-server transmission, and possible data reduction time. 
The model is supplemented with exhaustive profiling to obtain key parameters (e.g. chunk size which determines number of rounds of transmission for a message) optimized by CC libraries like \nccl for different hardware features and software configurations that affect the individual components above.
This hybrid approach strikes a good balance between accuracy and efficiency.
(3) It introduces an black-box ML-based slowdown predictor to model the slowdown caused by overlapping \ops. 
The XGBoost-based predictor relies on categorical features such as transmission protocol and channel configuration in \nccl, to numerical kernel-level performance statistics such as streaming multiprocessors (SM) occupancy and DRAM utilization, to capture their complex interactions.

We implement \sys to support mainstream training frameworks: PyTorch,  DeepSpeed~\cite{rasley2020deepspeed}, and Megatron-LM~\cite{shoeybi2019megatron}.
Our implementation with $\sim$10k LoC also includes a suite of automation tools to facilitate workload tracing and runtime profiling.
We evaluate \sys on H800 and A800 clusters across a variety of scenarios and models, achieving an average prediction accuracy of 91.4\% in training step time, and is up to 3x better than the state-of-the-art. 
For 96-GPU training of GPT-175B, \sys achieves 92\% accuracy in less than 2 minutes. 
We plan to open source \sys to the community.

\section{Background and Motivation} 
\label{sec:motivation}

\fyc{
\subsection{Large-scale Distributed DNN Training} 
\label{subsec:background}

\begin{figure}[t]
    \centering 
    \includegraphics[width=0.9\linewidth]{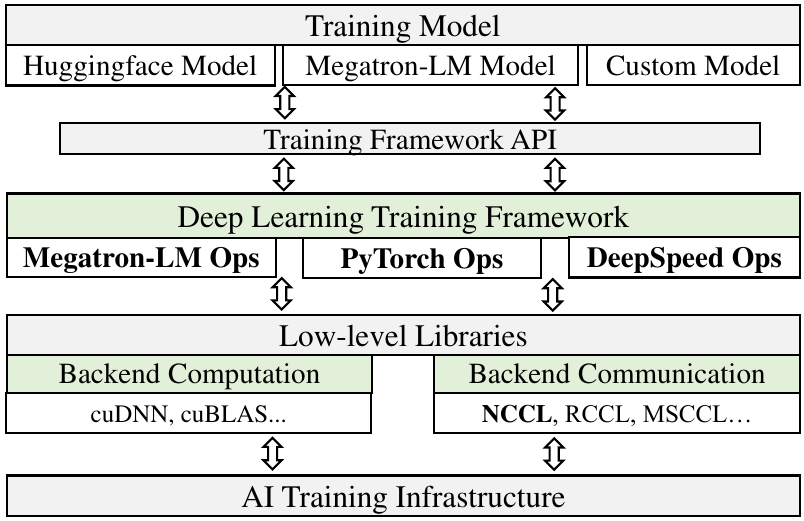} 
    \caption{Overview of the technical stack and architecture for large-scale model training.} 
    \label{fig:tech_overview} 
\end{figure}

Figure~\ref{fig:tech_overview} illustrates the layered technical stack underlying large-scale training, from high-level models and frameworks to low-level GPU libraries and networked systems. Today's state-of-the-art models often exceed 100 billion parameters~\cite{jiang2024preventing,llama3-meta}, and recent GPU architectures (e.g., NVIDIA Hopper and Ampere) combined with high-speed interconnects (e.g., InfiniBand and RoCE) enable efficient training at the scale of thousands of GPUs. New parallelization techniques—such as 3D parallelism~\cite{shoeybi2019megatron}, communication-computation overlap~\cite{wang2022overlap}, and ZeRO~\cite{rajbhandari2020zero}—further refine resource utilization, accelerating the convergence of these massive models.

\noindent\textbf{ML frameworks.} Modern large-scale training frameworks, such as Megatron-LM~\cite{shoeybi2019megatron} and DeepSpeed~\cite{rasley2020deepspeed}, have become de facto standards in production settings at companies like Microsoft, NVIDIA, and Alibaba~\cite{rasley2020deepspeed,shoeybi2019megatron,SimAI2025NSDI}. These frameworks implement 3D parallelism—combining DP, TP, and PP—to efficiently scale model training across large GPU clusters. In brief, DP replicates the model and splits the data, TP divides model parameters among GPUs, and PP stages model layers for pipelined execution. Together, these methods accelerate training of hundred-billion-parameter models. Furthermore, kernel-level overlap techniques (e.g., communication-computation overlap) and optimized kernels further minimize idle times and improve scaling efficiency.

\begin{table*}[t]
    \centering
    \small
    \setlength{\tabcolsep}{2pt}
    \renewcommand{\arraystretch}{1.0} 
    \begin{tabularx}{\textwidth}{
        >{\centering\arraybackslash}p{80pt} %
        >{\centering\arraybackslash}p{85pt} %
        >{\centering\arraybackslash}p{80pt} %
        >{\centering\arraybackslash}p{85pt} %
        >{\centering\arraybackslash}p{75pt} %
        >{\centering\arraybackslash}p{75pt} %
    }
    \toprule
    \multirow{2}{*}{Simulator} & 
    \multicolumn{2}{c}{Training Workload} & 
    \multirow{2}{*}{Communication} & 
    \multirow{2}{*}{3D Parallelism} & 
    \multirow{2}{*}{Overlap Slowdown} \\ 
    \cmidrule(lr){2-3}
    & Framework-specific op & \Ep & & & \\ 
    \midrule
    FlexFlow~\cite{jia2019beyond} & \xmark & \checkmark & ${\alpha-\beta}$ model & \checkmark & \xmark \\ 
    Daydream~\cite{zhu2020daydream} & \checkmark & \xmark & ${\alpha-\beta}$ model & \xmark & \xmark \\
    dPRO~\cite{hu2022dpro} & \checkmark & \xmark & ${\alpha-\beta}$ model & \xmark & \xmark \\ 
    DistSim~\cite{lu2023distsim} & \checkmark & \xmark & ${\alpha-\beta}$ model & \checkmark & \xmark \\ 
    Astra-sim~\cite{won2023astra} & \xmark & \xmark & ${\alpha-\beta}$ model & -- & \xmark \\ 
    Proteus~\cite{duan2024proteus} & \xmark & \checkmark & ${\alpha-\beta}$ model & \checkmark & -- \\ 
    SimAI~\cite{SimAI2025NSDI} & \checkmark & --  & Event-driven & -- & \xmark \\ 
    Echo (ours) & \checkmark & \checkmark & White-box modeling & \checkmark & \checkmark \\ 
    \bottomrule
    \end{tabularx}
    \vspace{0mm}
    \caption{Comparison of key features across simulators. \checkmark\ indicates full support, \xmark\ indicates no support, and \textbf{--} denotes partial or conditional support. Simulators are evaluated on training workload generation (framework-specific operations, ex-situ simulation), communication modeling, support for 3D parallelism, and handling of overlap-induced slowdowns. }
    \label{tab:work_compare}
\end{table*}

\noindent\textbf{Communication library.} Collective Communication Libraries (CCLs) facilitate efficient data transfer across GPUs in distributed systems~\cite{weingram2023xccl}. Among these, NCCL is the most widely adopted in production environments~\cite{weingram2023xccl,shoeybi2019megatron,SimAI2025NSDI,rasley2020deepspeed}, providing core communication primitives such as \ar, \ag, and \rs. NCCL serves as the default communication backend for mainstream training frameworks and incorporates numerous optimizations, including advanced topological algorithms and protocols that decompose data transmission into efficient point-to-point (P2P) transfers.
}

\subsection{Design Goals}

The overarching goal of this work is to build a practical simulator for large-scale distributed training.
Specifically, we aim to achieve the following goals: 

\begin{itemize}
    \item \textbf{High accuracy.} The basic goal is to accurately predict the  end-to-end training step time. 
    
    \item \textbf{\Ep.} 
    if the simulator's input is collected by deploying the job to the target cluster at full scale, i.e. \textit{in-situ} simulation, its utility is fundamentally limited by the available resources. 
    To be able to explore scales and scenarios beyond what is currently available, we desire \textit{\ep} design that can simulate a 1k-GPU cluster using a single machine for example.

    \item \textbf{High efficiency.} The simulator should be fast with minimal computational overhead especially for large-scale settings.
     
    \item \textbf{Usability.} The simulator should support mainstream training frameworks, models, and parallelism strategies. It must also be easy to use, automating the process as much as possible to minimize manual effort.

\end{itemize}

\subsection{Challenges} 
\label{subsec:challenges}

Training simulation has started to attract attention recently~\cite{SimAI2025NSDI,duan2024proteus,won2023astra,lu2023distsim,hu2022dpro}.
Inspired by prior work, we also separately consider computation and communication, and computation can be relatively easily simulated by offline profiling on a single device.
Yet, to achieve the four design goals, we identify three key challenges that prior work has not fully addressed.
Table~\ref{tab:work_compare} summarizes the comparison between \sys and existing simulators.  

Before we discuss the challenges, we note that all experiments in this section are done on a NVIDIA A100 GPU cluster.
Each 8-GPU node in the cluster has 600 GB/s NVLINK intra-node bandwidth, and four Mellanox ConnectX-6 NICs each with 2x100 Gbps. 
Training and bus bandwidth tests are performed using PyTorch's DDP and NCCL-test~\cite{nccltest}, respectively.

\begin{figure*}[t] 
    \centering 
    \includegraphics[width=1\linewidth]{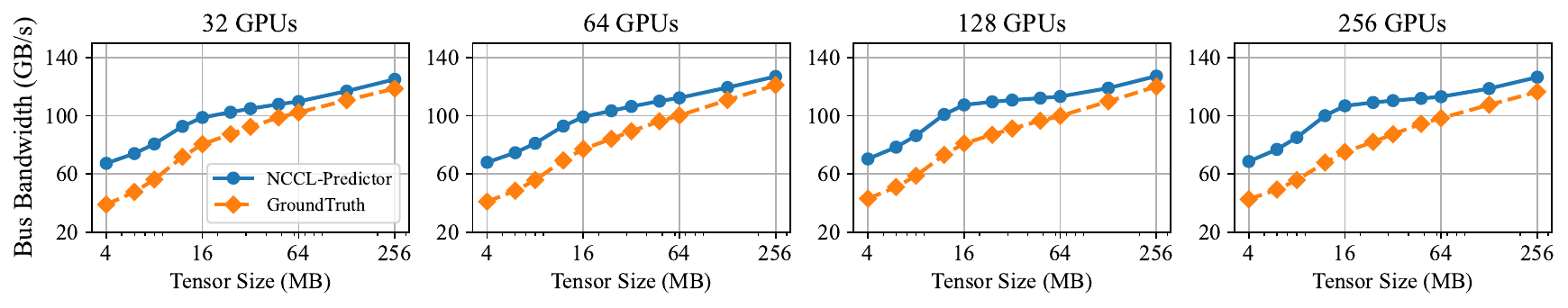} 
    \vspace{-3mm}
    \caption{Bus bandwidth of \ar communication operation with varying message sizes and GPU counts, profiled on A100 clusters.}
    \label{fig:bw_compare} 
\end{figure*}

\vspace{0.3em}
\noindent \textbf{Challenge 1:}
\textit{How to obtain the actual training workloads without a full-scale deployment?}

Accurate ML training simulation relies on the so-called training workload as the foundation. 
The \textit{training workload} refers to the execution graph that encodes the dependencies or execution sequence of computation and communication operators, plus other necessary information like tensor shapes and data types.
It serves as the blueprint for replaying the training process (on a given device). 
In large-scale 3D parallelism training, each GPU device (rank) independently initializes its assigned submodel based on TP and PP, resulting in unique workloads per rank.
These workloads are established only when the training framework begins execution.

Thus for accuracy, one naturally adopts a runtime tracing-based approach to capture the training workloads~\cite{hu2022dpro,lu2023distsim,luo2022srifty,zhu2020daydream}: the per-device execution graph is extracted after each device builds its execution graph in parallel. 
Yet this assumes a full-scale deployment of the job on the cluster, which severely limits the practical utility of the simulator in large-scale settings and is exactly one of the main motivations for building a simulator. 
Some systems like Proteus~\cite{duan2024proteus} and ASTRA-sim~\cite{won2023astra} choose to manually distribute and maintain the operators and generate the training workloads in order to avoid this limitation.
However, this hand-crafted approach is inherently imprecise since it fails to accurately represent the nuanced runtime characteristics of \textit{framework-specific \ops}—custom or highly optimized \ops tightly integrated with a particular framework's execution model (see \cref{subsec:workload_perf}). Such \ops often rely on environment-specific optimizations that emerge only under full-scale deployment conditions. Without capturing these contextual factors, manually approximating \op behavior becomes guesswork, leading to substantial inaccuracies.
For instance, we observed that ASTRA-sim overestimated computation \op by over 100\% on A800 GPUs when running Megatron-LM workloads.
How to faithfully capture the training workloads without a full-scale deployment becomes our first challenge.

\vspace{0.3em}
\noindent \textbf{Challenge 2:}
\textit{How to accurately simulate collective communication without high overheads?}

Communication \ops are critical in distributed training, yet hard to simulate as they involve complex interactions among devices and machines, and are affected by a range of network factors such as topology, latency, congestion control, etc. Existing approaches fall into two categories. 
Most simulators like ASTRA-sim~\cite{won2023astra}, FlexFlow~\cite{jia2019beyond}, Proteus~\cite{duan2024proteus}, and NCCL-Predictor~\cite{nccl} adopt a ${\alpha-\beta}$ model~\cite{valiant1990bridging}, which essentially calculates the running time of a communication \op as $tensor\_size/bandwidth + latency$.
Figure~\ref{fig:bw_compare} shows the most advanced NCCL-Predictor greatly overestimates the effective bus bandwidth of \ar \ops especially for smaller messages.
Note NCCL-Predictor already uses additional parameters to account for the impact of hardware, communication protocols (TCP vs RDMA), and algorithms (ring vs tree for \ar), but the over-simplified ${\alpha-\beta}$ model is inherently limited in capturing complex interactions across all these factors.
Moreover, NCCL optimizes distributed communication by integrating hardware offloading and acceleration techniques, such as GPUDirect RDMA~\cite{gdrdma} and GDRCopy~\cite{gdrcopy}, which lead to variations in internal kernel behavior. 

The second, more precise approach involves white-box modeling specifically targeting collective communication \cite{won2023astra,SimAI2025NSDI}.
Since \nccl (and other CCLs) implements a communication \op as an orchestrated series of P2P send and receive operations, we can simulate each send and receive to compose the end-to-end result. 
Similar to the training workloads, this naturally reflects the complex optimizations used by CCLs according to topology, NCCL protocol (LL, LL128 or Simple), network interconnect technologies (IB or RoCE), algorithm (ring or tree for \ar), etc.
To faithfully simulate the send/receive time, prior work \cite{won2023astra,SimAI2025NSDI} uses packet-level discrete-event network simulators like ns-3 \cite{riley2010ns}, which takes the actual topology and profiled link bandwidth as input.
Yet packet-level simulation is computationally expensive as it needs to simulate the entire protocol stack at each node in the network: 
We run the most recent SimAI \cite{SimAI2025NSDI} with an optimized multi-thread ns-3, and observe that simulating a single iteration on a 128-GPU cluster takes over 2 hours (7655s) on a 32-core server.
Therefore, the challenge is how to scale the white-box communication simulation efficiently without sacrificing accuracy.

\vspace{0.3em}
\noindent \textbf{Challenge 3:}
\textit{How to account for the interference between overlapping computation and communication?}

\begin{table}[t]
    \scriptsize
    \centering 
    \resizebox{1\columnwidth}{!}{ 
        \begin{tabular}{@{}lll|ll@{}}
        \toprule
        Model       & Overlapped op (\%) & Slowdown & Kernel Type & Slowdown \\ \midrule
        BERT\_Large & 59.2\% & 1.391    & Transform  & 1.142    \\
        GPT2        & 59.5\% & 1.481    & Sum    & 1.508    \\
        VGG19       & 57.5\% & 1.377    & GEMM   & 1.269    \\
        ResNet152   & 57.9\% & 1.435    & Grad   & 1.613    \\ \bottomrule
        \end{tabular}} 
    \caption{Statistics about (1) the proportion of overlapped kernels in different models and the slowdown factors in training step time, and (2) slowdown factors of some common computation \ops. Training is done on a single 8-GPU node of our A100 cluster with other settings unchanged; \ar is the only communication \op here in PyTorch DDP.} 
    \label{table:model_kernel_slowdown} 
\end{table}

\begin{figure}[t]
    \begin{minipage}{0.25\textwidth}
        \centering 
        \includegraphics[width=\linewidth]{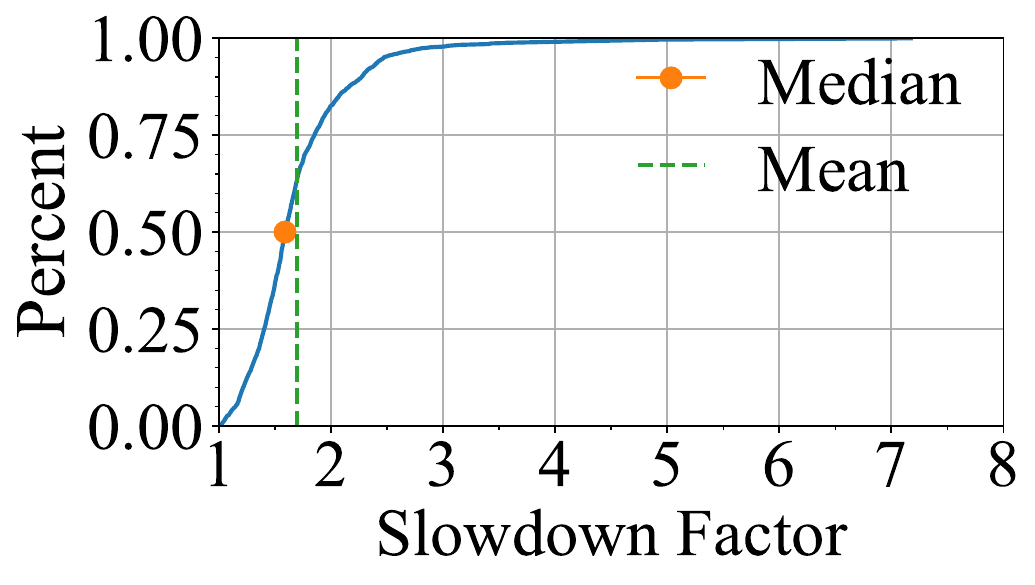} 
        \caption{CDF of slowdown factor of various kernels in GPT2.} \label{fig:slowdown_cdf} 
    \end{minipage}%
    \hfill
    \begin{minipage}{0.2\textwidth}
        \centering
        \small
        \resizebox{\linewidth}{!}{
            \begin{tabular}{@{}llcccc@{}}
            \toprule
            GPU & Kernel        & CUDA & Slowdown \\ \midrule
            \multirow{4}{*}{\centering 3090}         & GEMM       & 11.8  & 1.264     \\
                                                     & GEMM       & 12.1  & 1.393      \\
                                                     & LayerNorm  & 11.8  & 1.180 \\
                                                     & LayerNorm  & 12.1  & 1.938 \\ \midrule
            \multirow{4}{*}{\centering A800}         & GEMM       & 11.8  & 1.134 \\
                                                     & GEMM       & 12.1  & 1.178 \\
                                                     & LayerNorm  & 11.8  & 1.091 \\
                                                     & LayerNorm  & 12.1  & 1.207 \\ \bottomrule
            \end{tabular}}
        \captionof{table}{Performance comparison of CUDA kernels on different GPU architectures, showing execution slowdown across various CUDA versions.}
        \label{table:slowdown_env}
    \end{minipage}
\end{figure}

Overlapping computation and communication is widely used in practice to improve training efficiency and hardware utilization~\cite{jiang2024megascale,chen2024centauri,zhang2017poseidon}.
We find that overlapping introduces non-negligible slowdowns especially to computation \ops, even though they are assigned to separate streams on the GPU (contention for shared memory may be a possible cause).
Here, we define the slowdown factor as the ratio of the running time with overlapping communication kernels to the running time without communication kernels.
As shown in Table~\ref{table:model_kernel_slowdown}, more than 50\% computation \ops overlap with communication, leading to a slowdown of up to 1.48x in training step time.
We also examine the slowdown to individual kernels.
Table~\ref{table:model_kernel_slowdown} (second column group) shows when overlapped with \ar with 25MB messages, running times of common kernels increase by an average of 37.76\%.
The CDF of slowdown factors for all computation kernels of GPT-2 is presented in Figure~\ref{fig:slowdown_cdf}, where slowdown can reach 8 with an average of 1.70. 
Some recent work~\cite{hwang2023ark} also reported similar slowdowns in production training clusters. 
Unfortunately, this phenomenon has been overlooked by most existing simulation work despite its salient impact.
The only work we know that considers this is Proteus~\cite{duan2024proteus}: it simply uses a heuristic factor that varies only with GPU architecture and ML model to obtain the slowdown, which is too coarse to be accurate and not generalizable to unseen models.

Simulating the interference-induced slowdown is a daunting task. 
We ideally would need to model the kernel-level behavior in order to precisely know how different kernels overlap.
For example, in the widely-used gradient overlap optimization~\cite{zhang2017poseidon}, the simulator must correctly replay the backpropagation computation kernels and schedule communication kernels at precise times to achieve accurate overlap.
Further, for a given computation kernel, its slowdown factor is influenced by various factors of hardware and software setup. 
Table~\ref{table:slowdown_env} shows the slowdown factors of two kernels on different GPUs and CUDA versions vary a lot.
This is because the internal implementation, which varies based on the factors above, ultimately determines how resources are scheduled and shared among concurrent kernels.
These proprietary details are extremely inaccessible for us, not to mention the complexity of modeling them.
Thus, how to practically simulate the effect of overlapping computation and communication remains a formidable challenge.

\section{Design Overview} 
\label{sec:overview} 

Building upon the aforementioned observations and motivations, 
we introduce \sys, a simulator tailored for large-scale distributed training.

\begin{figure}[t] 
    \centering 
    \includegraphics[width=\linewidth]{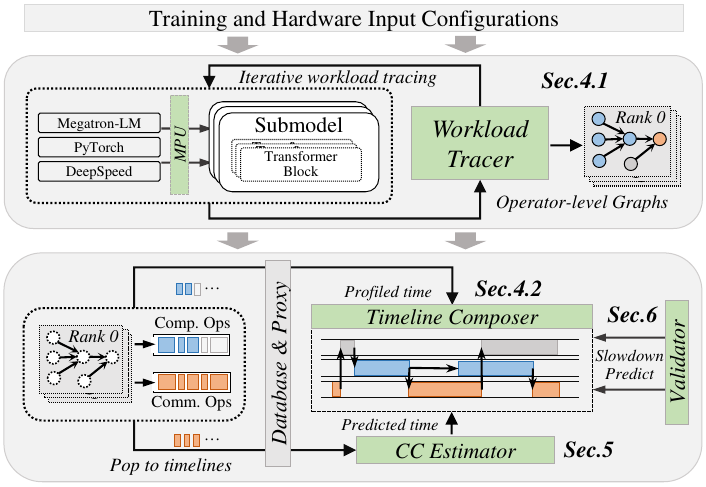} 
    \caption{\sys architecture overview. The green components represent the core modules of \sys.} 
    \label{fig:arch} 
\end{figure}

\noindent\textbf{Key design choices.}
We highlight the key choices in building \sys to address the three challenges in \cref{subsec:challenges}. 
\begin{itemize}[leftmargin=*]

\item To obtain training workloads without a full-scale deployment, \sys employs a novel ex-situ tracing approach. 
Generally, ML frameworks support various parallelism in distributed training in the following way: (1) they initialize each rank (typically one rank per device) with its own sub-model based on the parallelism setting, and then (2) each rank builds its own execution graph to start actual training. 
\sys turns this parallel process into a sequential one, using only one device to act as each rank iteratively to trace the corresponding execution graph and profile the running time of each \op at the same time, achieving faithful workload tracing without requiring the full cluster.

\item To achieve efficient communication simulation without high overheads, \sys employs a white-box modeling approach with empirically profiled parameters that strikes a good balance between accuracy and efficiency.
Our model is derived closely after NCCL's underlying chunk-based implementation of collective communication capturing critical factors such as per-chunk inter-server transmission time. 
These factors are profiled offline exhaustively to capture the complex dynamic optimization by \nccl according to hardware features and software configurations.

\item To account for the interference between overlapping computation and communication, \sys adopts a black-box method using features related to both computation and communication kernels, such as bucket size, SM occupancy, and cache hit rate.
\end{itemize}

\noindent\textbf{System architecture.}
Figure~\ref{fig:arch} shows \sys's architecture.
Its input includes three categories: user-defined settings specifying the training framework, parallelism strategies (DP/TP/PP groups); model details defining the model structure and hyperparameters; and hardware configurations detailing device specifics (GPU type), cluster size, network topology, \nccl parameters (e.g., NCCL\_TOPO\_FILE, NCCL\_BUFFSIZE), etc.
Given these inputs, the \textit{\extractor} extracts training workloads for each rank, including the per-rank execution graph and \op running times (\cref{subsec:workload_const}). 
The \textit{\cm} module (\cref{sec:comm_pred}) leverages the white-box models with parameters corresponding to the given settings to estimate each communication kernel's performance. 
Then the \textit{\ma} module (\cref{subsec:workload_replay}) re-constructs the global timelines of the end-to-end training process by assembling all per-rank execution graphs with estimated communication \op performance and inter-rank dependencies. 
The \textit{\va} module (\cref{sec:overlap_pred}) applies a machine learning model to adjust the running times of overlapping operations, accounting for the interference-induced slowdown, and outputs the final training step time result.  
Additionally, \sys maintains a \textit{database} for profiling data, reducing redundant measurements across simulations.

\section{Workload Tracing } 
\label{sec:workload_const_replay} 

We first present in this section how \sys accurately traces the training workloads for each rank in an ex-situ fashion using a single device, thereby addressing Challenge 1 (\cref{subsec:challenges}). 
We also show how \sys composes the global timelines across devices using the per-rank workloads to faithfully re-construct the entire training process in \cref{subsec:workload_replay}.

\subsection{Per-Rank Workload Tracing} 
\label{subsec:workload_const}

\begin{figure}[t] 
    \centering 
    \includegraphics[width=\linewidth]{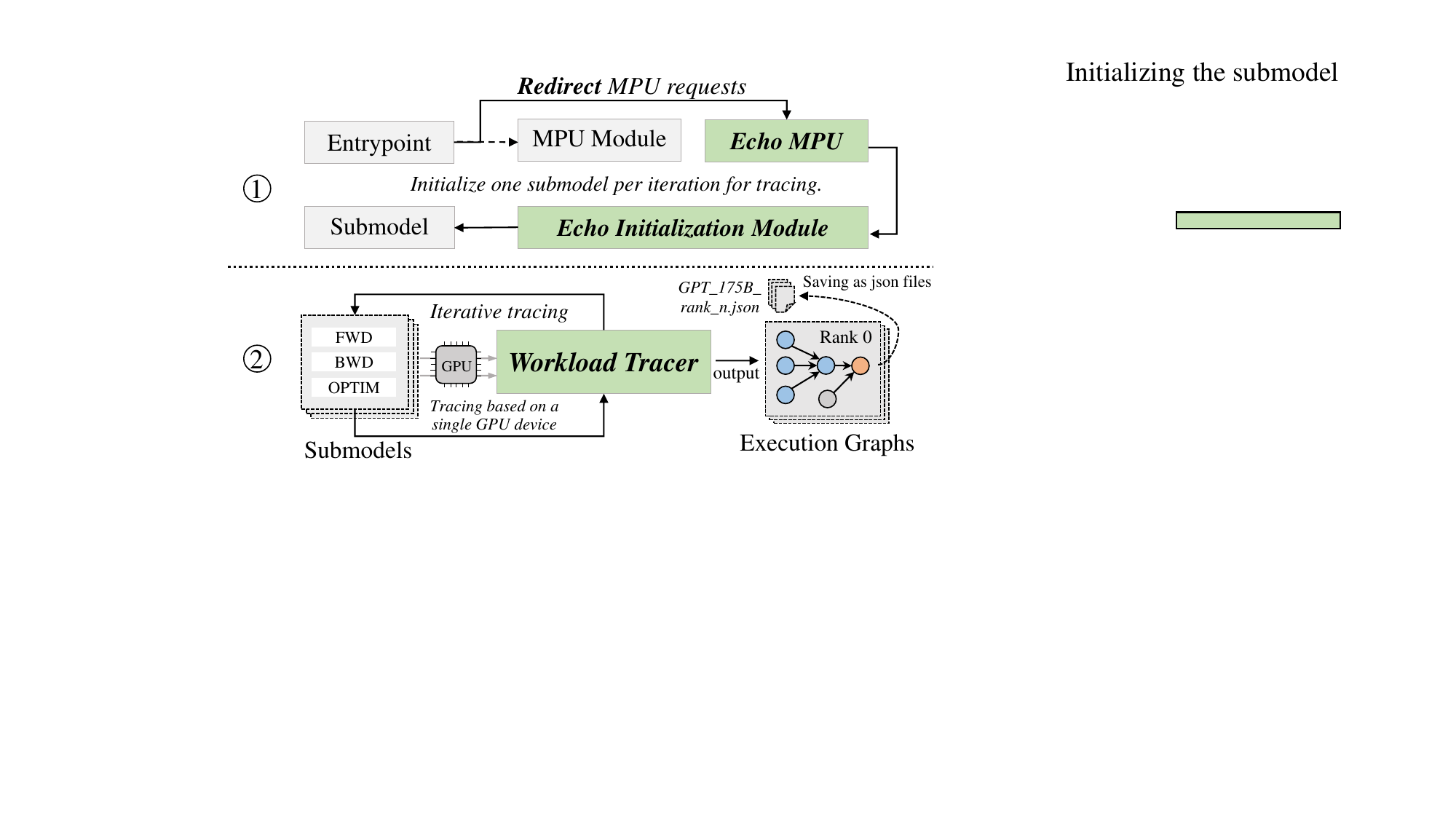} 
    \caption{\sys's MPU module and workload tracer.} 
    \label{fig:profier} 
\end{figure}

To launch a training job, the ML framework registers the global parallelism parameters (e.g. DP/TP/PP groups) given by user into the so-called model parallelism unit (MPU), which maintains all states related to parallelisms for a given rank.
Then according to the MPU, each rank concurrently initializes its own part of the model that it needs to train for (Figure \ref{fig:profier} \textcircled{1}
), and starts training with an execution graph that is optimized for its submodel based on the hardware/software environment and user configurations. 

\sys extracts the per-rank execution graph in an ex-situ fashion by transforming the above parallel process into a sequential one (Figure \ref{fig:profier} \textcircled{2}
). 
Specifically, \sys hijacks all the calls to the original MPU and re-directs them to its own MPU, and registers the global parallelism parameters as usual.
\sys's MPU only manipulates the input arguments to the MPU when necessary; it does not change the original implementation to ensure correctness.
Then with a single device, in each iteration $i$, \sys uses this unique rank ID $i$ to initialize the corresponding submodel and execute one complete training step, so all information regarding the computation \ops, including their dependencies (including those w.r.t other ranks) and running times, can be profiled.

The caveat here is collective communication: one device obviously cannot execute a communication \op. 
To resolve this, notice that when the communication \op is launched it also needs to call into the MPU which determines the device's position in the global communication groups for proper execution.
This is re-directed to \sys's MPU again allowing it to trace the communication \op's truthful information in the simulated setting (group association, message sizes, etc.); meanwhile, instead of returning results using the simulated (distributed) setting, \sys's MPU returns results corresponding to single device setting, so the communication \op can proceed and training is not blocked. 
This way \sys traces the actual execution graphs for each rank and running times of computation \ops without requiring the full-scale cluster.
Implementation details of tracing vary across frameworks which we will discuss in \cref{sec:implementation}.

Note that after per-rank workload tracing, all collective communication \ops on the per-node execution graph are placeholders without any running time. 
The job of \sys's \cm and \va is precisely to estimate and refine the communication times, respectively, in the next phase.

\subsection{Timeline Composing } 
\label{subsec:workload_replay}

Before presenting the communication estimation and validation, we discuss how \sys obtains the end-to-end training step time by composing the global timeline with the per-rank workloads obtained above.

\begin{figure}[t] 
    \centering 
    \includegraphics[width=\linewidth]{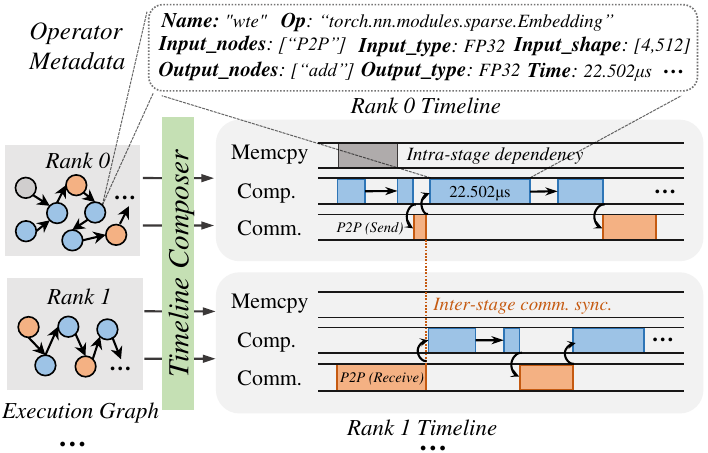} 
    \caption{Illustration of how \sys schedules operators in the execution graph to timelines across ranks.} 
    \label{fig:timeline} 
\end{figure}

Assuming all communication running times are known, \sys's \ma can recover the training process by replaying the per-rank execution graphs.
\ma is an event-driven simulator that walks through the per-rank execution graphs and puts all \ops (events) onto the global timelines accordingly, including the computation timeline, communication timeline, and memory copy (memcpy) timeline. 
As shown in Figure~\ref{fig:timeline}, a critical piece of information missing from the per-rank graphs is dependencies across ranks; for instance the downstream stages in rank $i$ requires activations from the upstream stages in rank $i-1$ before starting the forward computation in PP.
We provide predefined dependencies and matching rules for both inter-stage and intra-stage events for common parallelism strategies, such as 1F1B.
To enhance flexibility, this module is exposed as an API, allowing users to input new dependencies and matching rules for custom strategies.

\section{Communication Estimation} 
\label{sec:comm_pred} 

In this section, we outline our approach to simulate the running time of communication \ops. 
We provide some key background on \nccl in \cref{subsec:tree_allreduce} as the basis of our approach.
Then we present our white-box modeling in \cref{subsec:whitebox_modeling} and \cref{subsec:allreduce_formulation} for synchronization and execution time of a communication kernel, respectively, in addressing Challenge 2 of \cref{subsec:challenges}.

\subsection{Background on \nccl} 
\label{subsec:tree_allreduce}
NCCL provides collective communication (CC) functions, such as \ar, \ag, and \rs, to support various parallelisms in training. 
These operations have a variety of implementations optimized for different algorithms, protocols, etc. 

In NCCL, P2P primitives and CC functions are implemented within a single CUDA kernel \cite{nccl}. 
Thus, our modeling and prediction here is done directly at the kernel level. 
The kernel-level results are used as the corresponding \op's performance.
This results in inaccuracies since the launching overhead of the \op is not considered, which however is very small.
The same applies to the slowdown prediction in \cref{sec:overlap_pred}.

Each message is divided into equal-sized chunks that are pipelined for efficient synchronization and transfer.
This is the basis of our modeling in \cref{subsec:allreduce_formulation}.

\begin{figure}[t] 
    \centering 
    \includegraphics[width=1\linewidth]{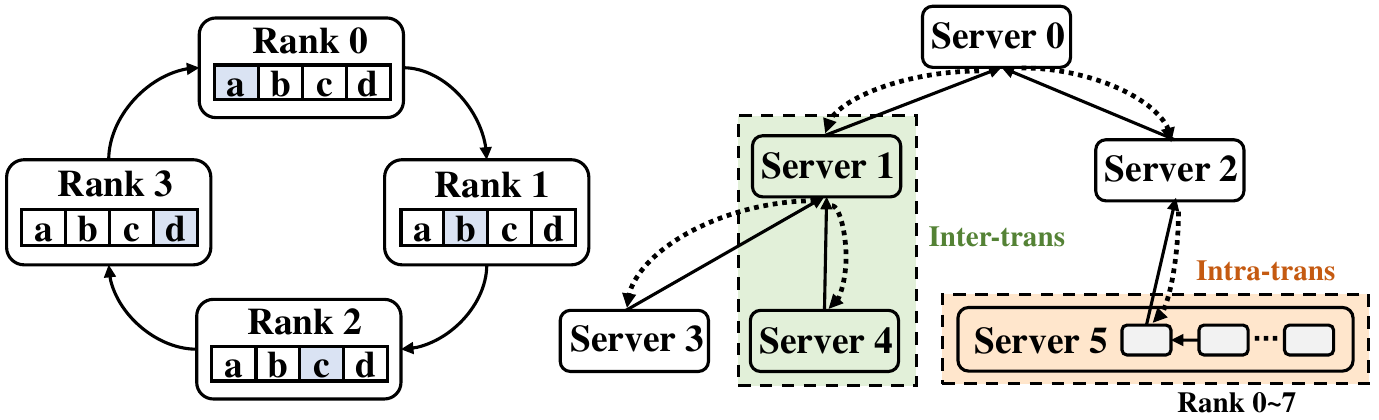} 
    \vspace{-4mm}
    \caption{Illustration of ring-based (left) and tree-based (right) communication execution. In the tree-based communication, black solid arrows denote the reduce process, while dashed arrows indicate the broadcast process.} 
    \label{fig:ring_tree_logic} 
\end{figure}

NCCL supports two common algorithms for \ar: ring-based~\cite{nccl} and tree-based~\cite{tree}. 
Figure~\ref{fig:ring_tree_logic} illustrates how they work.
Other CC kernels use ring-based algorithm only.

\subsection{White-Box Modeling: Synchronization}
\label{subsec:whitebox_modeling}

\sys models a \nccl kernel as two sequential stages: synchronization and execution. Upon invocation, the kernel first checks whether it can start execution based on the readiness of ranks in the current communication group. If not all ranks are ready, the kernel waits for synchronization before it can proceed. 
Thus, the total running time of a kernel is the sum of synchronization time and execution time.
Here we investigate the synchronization time first. 

\sys models synchronization time after NCCL's implementation logic.
For P2P \send and \recv, it waits until its target communicator (i.e., each rank's communicator instance) is ready to proceed. 
In contrast, for CC kernels, the communicator associated with each rank in the current group must wait until all communicators have been successfully launched. 
The last communicator to initiate determines the actual start time of this kernel.

\subsection{White-Box Modeling: Execution}
\label{subsec:allreduce_formulation}

Execution time represents the majority of kernel running time, and is the focus of our modeling.
We build white-box models to closely model the kernel execution process in NCCL with different algorithms, and profile all the key parameters of the models exhaustively in various hardware/software settings to achieve accurate prediction without high overheads of event-based simulation.

\noindent\textbf{Basic model.}
Based on NCCL's GPU-side implementation, the execution time of a CC kernel can in general be broken down into four non-overlapping components, using \ar as the example:
\begin{equation} 
    \begin{split} 
        T_{comm\_kernel} = & T_{conn\_setup} +T_{intra\_trans} 
        \\ & +T_{data\_reduction} +T_{inter\_trans}, 
    \end{split} 
    \label{eqn:allreduce_components} 
\end{equation}
where $T_{conn\_setup}$ is the connection setup time; $T_{intra\_trans}$ and $T_{inter\_trans}$ are times to transmit the tensor chunks in the specific algorithm, $T_{data\_reduction}$ is the computation time of the reduce operation over the received tensor chunks with the local tensor.
Not all components are present in other CC kernels.

Based on Equation~\eqref{eqn:allreduce_components}, we can formulate the following for all CC kernels with different algorithms: 
\begin{align}
    T^{Ring}_{all-gather} = &\ \alpha + \gamma\cdot \eta (N-1), \label{eqn:44}\\  
    T^{Ring}_{reduce-scatter} = &\ \alpha +  \eta\big[(N-1) + \gamma \times (N-1)\big]\nonumber\\
                            &  + \delta \times tensor\_size, \label{eqn:33}  \\
    T^{Ring}_{all-reduce} =&\ \alpha + \eta\big[(N-1) + \gamma \times 2(N-1)\big]\nonumber\\
                        &\ + \delta \times tensor\_size , \label{eqn:11} \\
    T^{Tree}_{all-reduce} =&\ \alpha + \gamma (\eta-1) +  2\beta(K-1) \nonumber\\
                    &\ + 2\gamma\log_2(M) + \delta \times tensor\_size. \label{eqn:22}
\end{align}
Here $N$ is the total number of devices, $M$ number of nodes (servers), and $K=N/M$ number of devices per node. 
We denote the connection setup time as $\alpha$, intra-server and inter-server transmission times as $\beta$ and $\gamma$.
The reduce time depends on $tensor\_size$, and the reduce throughput $\delta$.
Finally, $\eta$ is the number of rounds or number of chunks for this tensor, which corresponds to \nccl's chunk-based implementation.

\noindent\textbf{Offline exhaustive profiling.} 
The modeling above is explicitly related to nine parameters and implicitly related to an additional parameter, $chunk\_size$. 
Among them, $N$, $M$, $K$ and $tensor\_size$ are user configurations while $\alpha, \beta, \gamma, \delta$, $\eta$ are what \sys attempts to obtain. However, modeling them explicitly is challenging. 
First, they obviously depend on $chunk\_size$ and $tensor\_size$. 
The chunk size is not a predefined constant but dynamically tuned by NCCL~\cite{nccl}. 
Further, they also vary with hardware configuration and software implementation (e.g. IB vs NVLink, \nccl version). 
These idiosyncrasies also vary across different CCLs, making it even more difficult to have a general model.  

We adopt offline profiling instead to obtain the values of these parameters, since they are fixed when all the hardware/software configurations are settled. 
First, we modify \nccl to be able to obtain the chunk size during the initialization phase. 
We enumerate all possible combinations of $N$, $M$, $tensor\_size$, and hardware type (IB, TCP, NVLink) to record the corresponding $chunk\_size$ and number of rounds $\eta$ computed by \nccl. 
Then for each $chunk\_size$, we profile the corresponding connection setup time, intra-server and inter-server transmission time, and reduce throughput ($\alpha, \beta, \gamma, \delta$) under the corresponding setting.
This does require access to a cluster as the network bandwidth may vary depending on the scale (2-node vs 4-node vs 8-node), but for common Clos-based cluster topologies they remain stable after reaching a small scale due to the symmetry and limited number of tiers. 

Taken together, our approach of combining white-box modeling with offline profiled parameters provides a good first-order estimation of CC kernel performance with very low overheads.
Our approach can be enhanced in several ways, which we discuss in \cref{sec:discussion}.

\section{Overlapping-Induced Slowdown Prediction} 
\label{sec:overlap_pred} 
Accurately capturing and modeling the resource contention between overlapping kernels presents substantial challenges in predicting the slowdown factor as discussed in \cref{subsec:challenges}. 
To address this, \sys introduces an ML-based slowdown prediction model with a number of hardware- and software-specific features, which we present here.

A GPU devices executes concurrent computation and communication kernels with spatial multitasking, where each kernel is assigned to separate streams with a subset of the SMs.
At the same time they also contend for shared resources such as cache and DRAM bandwidth.
These factors results in performance interference, and \sys needs to capture them in predicting the slowdown.

\begin{table}[t] 
    \centering 
    \resizebox{\columnwidth}{!}{ 
        \begin{tabular}{@{}cll@{}} 
            \toprule \multicolumn{1}{l}{Type} & Feature & Specification \\ \midrule 
            \multirow{6}{*}{\begin{tabular}[c]{@{}c@{}}Comm. \\ details \end{tabular}}
            & Protocol & Communication protocol used within NCCL \\ 
            & Algorithm & Communication topology algorithm applied in NCCL \\ 
            & Collective & Name of the collective function in the kernel \\ 
            & Bucket size & Size of the tensor bucket used in collective communication \\ 
            & Channel number & Number of communication channels in NCCL \\ 
            \midrule \multirow{8}{*}{\begin{tabular}[c]{@{}@{}c@{}}Comp. \\ Kernel \\ Metric \end{tabular}} 
            & Running time &  Running time profiled on single-device training \\
            & Compute throughput & Average SM throughput\\ 
            & Memory throughput & Percentage of cycles where DRAM was active\\ 
            & DRAM throughput & Peak DRAM throughput \\ 
            & Achieved occupancy & Average percentage of active warps on each SM \\ 
            & L1 hit rate & Hit rate for L1 cache \\ 
            & L2 hit rate & Hit rate for L2 cache \\ 
            \bottomrule 
        \end{tabular}} 
    \caption{Features in \sys's prediction model for predicting interference. } 
    \label{table:features_interference} 
\end{table}

\noindent\textbf{Feature extraction.} 
Table~\ref{table:features_interference} summarizes the key features used by our model. We categorize features into two groups: communication-related details and computation kernel metrics. 
For communication, we consider features such as protocol and algorithm employed by NCCL, CC kernel type, bucket size, and channel configuration. These parameters capture the complexity and overheads of communication: bucket size, for instance, determines when a communication kernel will be launched after accumulating one bucket of data. 
On the computation side, we profile kernels at a fine-grained level on a single device to measure their baseline performance without overlapping. 
We collect running times, compute throughput, memory utilization, and cache statistics using NVIDIA Nsight Systems and Nsight Compute. 
For instance, achieved occupancy and SM activity levels indicate compute intensity, while DRAM utilization and memory throughput---along with L1/L2 hit rates---shed light on memory access latency and potential bottlenecks. Additionally, we classify kernels by their functional types (e.g., matrix multiplication, layer normalization) to further refine our predictions.

\noindent\textbf{XGBoost-based model.} We leverage XGBoost~\cite{xgboost}, a tree-based gradient boosting framework, to predict the slowdown factors caused by overlapping kernels with the above features. 
XGBoost is well-suited to this task for two key reasons. First, our dataset contains heterogeneous features that are both numerical (e.g., SM throughput, DRAM utilization) and categorical (e.g., GPU type, kernel type), a scenario where tree-based models excel. 
Second, our target slowdown factors, influenced by the interplay of diverse system conditions, remain bounded within a range well-captured by the training set. As we expand our dataset---incorporating additional profiles from various models, GPU architectures, and cluster scales---the model's generalization ability improves, enabling accurate slowdown predictions across increasingly complex and heterogeneous environments.

\section{Implementation}
\label{sec:implementation}

We implement \sys as illustrated in Figure~\ref{fig:arch}. 
\sys is developed based on PyTorch 2.1, DeepSpeed 0.13.1, and Megatron-LM (commit ID: 53a350ed), and NCCL 2.22.3, comprising $\sim$10K lines of code (LoC). 
$\sim$2K LoC are dedicated to Megatron-LM, 3K LoC to DeepSpeed and PyTorch, and the remaining 5K LoC for common core modules that underpin the simulation engine, training workload tracing, overlapping induced slowdown dataset collection, and CC parameter exhaustive profiling.

\noindent \textbf{Workload tracer.} 
We implement two tracing modules to support \wg extraction for Huggingface models used by PyTorch and DeepSpeed and Megatron-LM models. For the former, the tracer leverages \texttt{torch.fx} to record \ops. 
Since \texttt{torch.fx} only captures computation \op during the forward pass, we enhance it to capture \ops during backward passes by using tensor gradient functions and registering hooks, as well as optimizer steps and communication \ops (handled as placeholders).
For Megatron-LM models, a custom tracing module is developed with a series of Python decorators and context managers. \sys directs the handling of custom \ops in Megatron-LM to a general template, ensuring that the related function calls are recorded. Finally, the generated \wg is output as a JSON file and stored in a database for future use. 
Notably, users can activate \sys's \extractor by adding a few extra lines of in their existing training scripts.

\noindent \textbf{\cm.} 
To profile the key parameters for models in \cref{subsec:whitebox_modeling}, we utilize the NCCL profiling tool NPKit~\cite{NPKit}.
Given that the network topology can impact bandwidth, we empirically profile these parameters for settings up to 4 nodes (32 GPUs) in our clusters, beyond which the effective per-node network bandwidth does not change (and thus these parameters).

\noindent \textbf{Validator/Slowdown predictor.}
We experimentally find the best hyperparameters for our XGBoost model as shown in Table~\ref{tab:hyperparameters}.
We implement an automated pipeline to construct the training dataset. 
NVIDIA Nsight Compute is used to profile GPU resource metrics such as SM throughput, memory bandwidth, and cache hit rates. 
Nsight Systems capture kernel execution traces under overlapping and non-overlapping conditions, exporting the data as SQLite databases.

\begin{table}[t]
    \scriptsize
    \centering
    \begin{tabular}{|c|c|}
        \hline
        \textbf{Hyperparameter} & \textbf{Value} \\
        \hline
        Model & XGBRegressor \\
        \hline
        Objective & reg:squarederror \\
        \hline
        Max depth & 12 \\
        \hline
        Random state & 42 \\
        \hline
        Train/Test Ratio & 0.8/0.2 \\
        \hline
        K-fold cross-validation & 5-fold \\
        \hline
    \end{tabular}
    \caption{The hyperparameters of XGBoost model.}
    \vspace{-3mm}
    \label{tab:hyperparameters}
\end{table}

\section{Evaluation} 
\label{sec:evaluation} 
This section evaluates \sys on vision and language models using two clusters up to 96 GPUs. We describe the experimental setup in \cref{subsec:setup}, then assess \sys's computation and communication simulation accuracy against baselines in \cref{subsec:workload_perf} and \cref{subsec:allreduce_perf}, respectively. We present \sys's kernel overlap slowdown prediction results in \cref{subsec:interference_perf}. Finally, we demonstrate its end-to-end simulation accuracy and time cost in \cref{subsec:overall_perf}.

\subsection{Experiment Setup} 
\label{subsec:setup}

\noindent\textbf{Models.} 
We evaluate \sys using two representative models: the VGG19~\cite{simonyan2014very} image classification model and GPT-series language models~\cite{shoeybi2019megatron}.
For GPT-series, we consider a broad range of model sizes (from 13B to 175B parameters; see~\cref{sec:model_configurations} for details) to thoroughly assess the simulator's accuracy.
To avoid precision-related discrepancies with the baselines, we use FP32 during evaluation.
For ground-truth measurements, we train the models on real GPU clusters using mainstream frameworks (PyTorch, DeepSpeed, Megatron-LM) with CUDA 12.1 aligned with the implementations in \cref{sec:implementation}.
For each measurement, we start profiling from the 5th training step, warm up for two steps, and then collect running times for the next five steps. We report the average of these five steps as the ground truth.

\noindent\textbf{Clusters.} 
To assess \sys under various hardware and topological conditions, we conduct experiments on two real-world GPU clusters.
The first is a large-scale cluster with 96 NVIDIA H800 GPUs (80GB each) distributed across 12 servers, where each server hosts 8 NVLink-connected GPUs (400 GB/s). Inter-server communication is provided by 400 Gb/s NDR IB per GPU. 
Each server is equipped with dual 56-core Intel Xeon Scalable processors and PCIe Gen5.
The second is a single-server cluster with 8 NVIDIA A800 GPUs connected via 400 GB/s NVLink.
Additionally, we include an RTX 3090 cluster in the kernel slowdown experiment (\cref{subsec:interference_perf}) to further diversify our hardware evaluation.
\Cref{sec:additional_obs} provides additional observations and evaluations on a 256-GPU A100 cluster.

\vspace{0.2em}
\noindent\textbf{Baselines.} 
We compare \sys with two fully open-source simulators:

\begin{itemize}[leftmargin=3mm]
    \vspace{-0.1em}
    \item \textbf{Proteus}~\cite{duan2024proteus}: A SOTA simulator that, like \sys, provides integrated simulation for parallel strategies and runtime behaviors.
    \vspace{-0.1em}
    \item \textbf{FlexFlow}~\cite{jia2019beyond}: An automated parallelization framework with an internal simulator for throughput estimation across different parallelization strategies.
\end{itemize}
\vspace{-0.1em}
Other simulators such as ASTRA-sim~\cite{won2023astra} and SimAI~\cite{SimAI2025NSDI} are also open source, but they have critical limitations. ASTRA-sim only supports matrix multiplication kernels; and SimAI's current open-source version (commit ID:44a468c as of Dec. 2024)~\cite{SimAI} {does not support pipeline parallelism}.

\begin{figure}[t] 
    \centering 
    \includegraphics[width=1\linewidth]{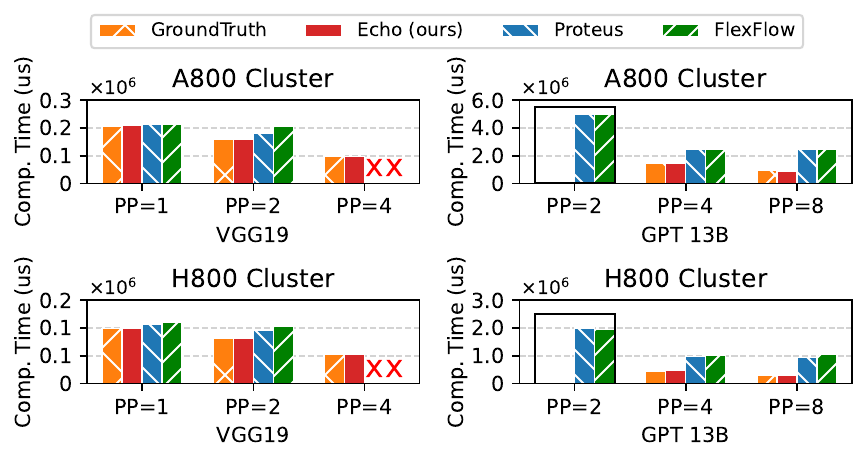} 
    \vspace{-5mm}
    \caption{
    Computation times of various workloads on A800 and H800 clusters under different parallelization settings. 
    Red crosses indicate configurations that \bs and \ff cannot support PP=4 for VGG19, 
    while black squares  mark configurations that encounter OOM errors at runtime.
    }
    \label{fig:comp_h}
\end{figure}

\subsection{Computation Simulation Performance}
\label{subsec:workload_perf}

\begin{figure}[t] 
    \centering 
    \includegraphics[width=1\linewidth]{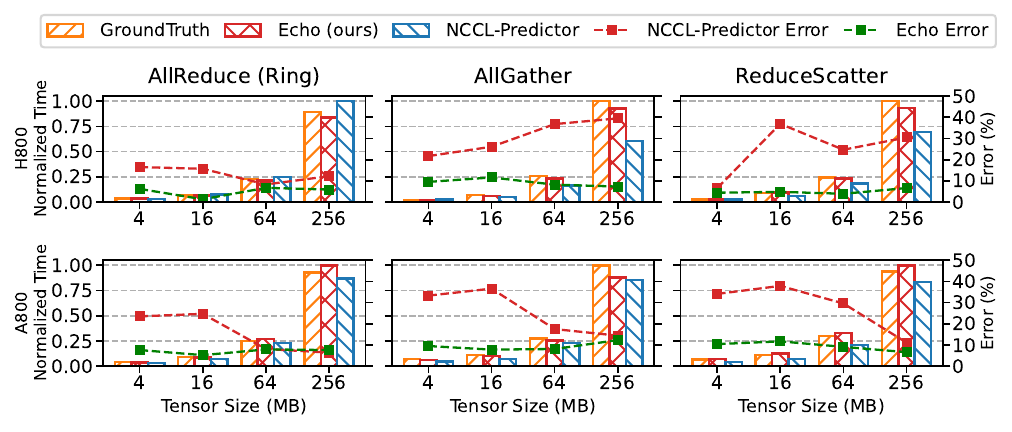} 
    \caption{Running time and prediction error for intra-server CC kernels on H800 and A800 GPUs across varying tensor sizes. Each subplot is normalized relative to the operation with the longest running time.}
    \label{fig:intra_nccl} 
\end{figure}
\begin{figure}[t] 
    \centering 
    \includegraphics[width=1\linewidth]{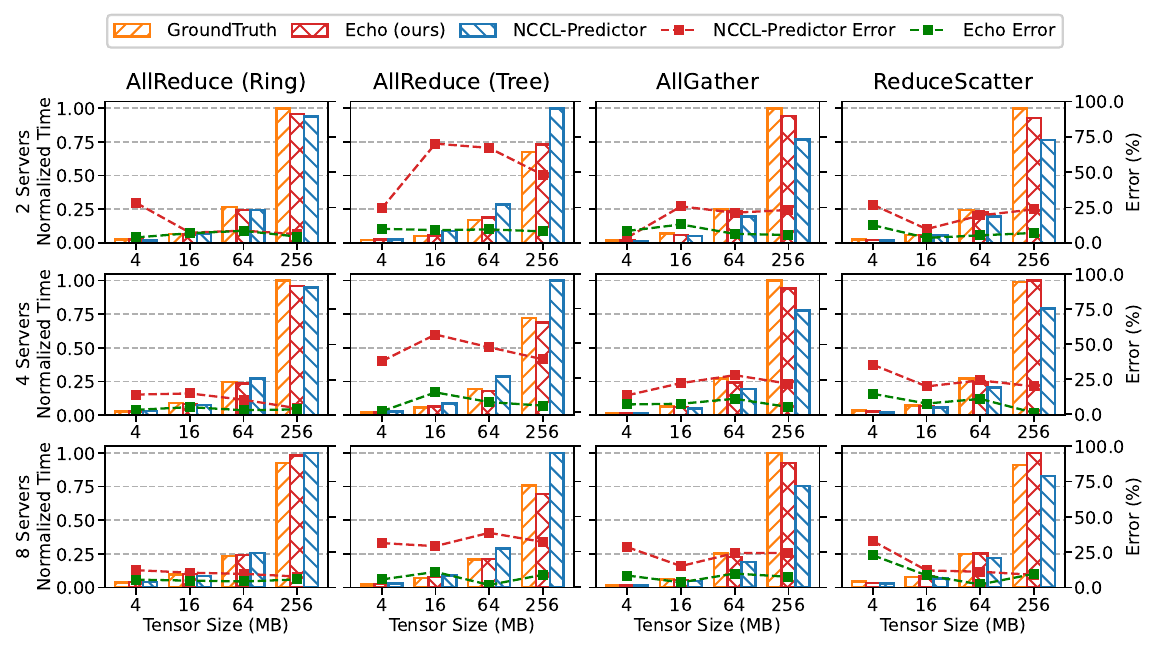} 
    \caption{Running time and prediction error for inter-server CC kernels on H800 GPUs across varying tensor sizes and cluster scales. Each subplot is normalized relative to the operation with the longest running time.}
    \label{fig:inter_nccl} 
\end{figure}

We first evaluate \sys's accuracy in simulating computation workloads on both H800 and A800 clusters, using VGG19 (with DeepSpeed) and GPT-13B (with Megatron-LM) under various PP configurations. Each configuration assigns a distinct submodel to an individual GPU.

\noindent\textbf{Simulation accuracy.} \sys reconstructs the computation workload by tracing all computational \ops and reproducing their execution sequence based on the \wg. Per-GPU \op running times are derived through \sys's \extractor without requiring a full-scale deployment. As shown in Figure~\ref{fig:comp_h}, \sys achieves a maximum overall error of only 8.31\%. In contrast, while \bs and \ff maintain relatively low error rates for VGG19 under DeepSpeed (8.6\% and 17.81\%, respectively), their errors escalate dramatically for GPT-13B under Megatron-LM, reaching 91.53\% and 109.14\%, respectively.
The root cause of this discrepancy is that Proteus and FlexFlow rely heavily on standard PyTorch \ops, failing to incorporate the custom fused \ops that frameworks like Megatron-LM employ for performance optimization. Such specialized \ops cannot be accurately modeled without extracting their runtime characteristics directly from the target training framework. By doing precisely this, \sys captures the actual \op behaviors and achieves substantially higher accuracy.

\sys also supports a wide range of 3D parallelism configurations. In contrast, Proteus and FlexFlow struggle with certain settings. For example, they cannot simulate VGG19 at PP=4 due to the need for manual workload construction. Finally, \sys is GPU memory-aware during workload extraction, it detects Out-of-Memory (OOM) conditions and provides immediate feedback. Neither Proteus nor FlexFlow anticipates OOM errors (e.g., GPT-13B at PP=2), denying users critical insights into the feasibility of configurations.

\subsection{Communication Simulation Performance}
\label{subsec:allreduce_perf}

We now evaluate the effectiveness of \sys's communication simulation, comparing it against NCCL-Predictor~\cite{nccl}, the underlying communication model used by both \bs and \ff. NCCL-Predictor leverages parameters tuned from production experience and employs an ${\alpha-\beta}$ model~(\cref{subsec:challenges}).

\noindent\textbf{Simulation accuracy.}
Figures~\ref{fig:intra_nccl} and~\ref{fig:inter_nccl} illustrate the prediction accuracy for intra- and inter-server collective communication under varying message sizes (4MB to 256MB) and cluster scales. Across most cases, \sys consistently outperforms NCCL-Predictor.
In the intra-server setting, \sys achieves an average prediction error of 8.43\% and 7.24\% on A800 and H800 clusters, respectively, compared to 26.75\% and 28.33\% for NCCL-Predictor. 
In the inter-server scenario, \sys's average prediction errors are 11.36\%, 12.59\%, and 13.75\% for 2-, 4-, and 8-server H800 clusters, respectively, outperforming NCCL-Predictor by a margin of 34.4\%, 29.56\%, and 19.39\%. The smallest prediction gap emerges in ring-based \ar, where \sys is only 4.2\% more accurate than NCCL-Predictor. By contrast, for other communication patterns, especially tree-based \ar, \sys demonstrates a more pronounced accuracy advantage, reducing average prediction errors by up to 33.5\%.

These differences can be attributed to NCCL's specialized optimizations for the widely used ring-based \ar, which align more closely with the assumptions of the ${\alpha-\beta}$ model. However, NCCL-Predictor's simplifications become problematic for less commonly optimized patterns and smaller message sizes, leading to significant discrepancies. In particular, NCCL-Predictor exhibits an average error rate of 27.6\% for 4MB and 16MB messages. By contrast, \sys's profiling-based white-box approach more accurately captures practical bandwidth utilization and overheads, ensuring consistent accuracy across both small and large message sizes. 
Notably, \sys's white-box communication prediction model significantly accelerates simulation speed, as discussed in~\cref{subsec:overall_perf}.

\begin{table}[t] 
    \centering
    \small
    \resizebox{\columnwidth}{!}{ 
        \begin{tabular}{@{}llllllll@{}}
            \toprule
            Tensor size (MB)           & 4      & 8       & 16    & 32      & 64      & 128     & 256     \\
            Chunk size ($10^3$ bytes)  & 18     & 36      & 72    & 72      & 72      & 144     & 288     \\ \midrule
            \begin{tabular}[c]{@{}l@{}}Intra-server \\ Round trip ($\mu$s) \end{tabular} & 57.43  & 64.36  & 73.62 & 74.61   & 77.46   & 75.55   & 99.23   \\ \midrule
            \begin{tabular}[c]{@{}l@{}}Inter-server \\ Round trip ($\mu$s) \end{tabular}   &  54.3 & 	72.32 & 	106.73 & 	112.9 & 	111.25	& 174.2	& 278.98 \\ \midrule
            Reduce operation ($\mu$s) &  50.14	& 65.86 &	102.46 &	219.89	& 463.43 &	841.71 &	1657.72  \\ \bottomrule
        \end{tabular}}
        
    \caption{The breakdown of \sys's \ar running time, profiled in the H800 cluster using 32 GPUs.}
    \label{table:allreduce_param}
\end{table}

\noindent\textbf{Breakdown analysis.}
Table~\ref{table:allreduce_param} provides a detailed breakdown of the \ar kernel runtime in a 32-GPU H800 cluster, including intra-server and inter-server round-trip times and reduce operation computation times for various tensor sizes. Intra-server round-trip times scale with tensor size, increasing from 57.43~\textmu s for 4MB tensors to 99.23~\textmu s for 256MB tensors, reflecting overheads such as scheduling more thread blocks for larger chunks. Similarly, inter-server round-trip times grow with tensor size, reaching 278.98~\textmu s at 256MB. The computation time for the reduce operation also scales nearly linearly with tensor size, increasing from 50.14~\textmu s at 4MB to 1657.72~\textmu s at 256MB. These trends highlight how \sys's detailed modeling captures realistic overheads across varying message sizes and communication settings.

\subsection{Slowdown Prediction} 
\label{subsec:interference_perf}

\begin{table}[t]
    \centering
    \small
    \resizebox{\columnwidth}{!}{
        \begin{tabular}{@{}llcccc@{}}
        \toprule
        Solution        & GPU Type & 5\% Error & 10\% Error & 15\% Error & RMSE \\ \midrule
        \sys (ours)     & \multirow{2}{*}{\centering 3090} & 61.48\% & 73.62\% & 76.52\% & 1.0935 \\
        Proteus         &                           & 8.52\%  & 13.62\% & 19.28\% & 1.4953 \\ \midrule
        \sys (ours)     & \multirow{2}{*}{\centering A800}     & 61.03\% & 51.59\% & 75.01\% & 0.7147 \\
        Proteus         &                           & 59.04\% & 65.45\% & 70.51\% & 1.5170 \\ \midrule
        \sys (ours)     & \multirow{2}{*}{\centering H800}     & 43.18\% & 51.59\% & 57.73\% & 0.5626 \\
        Proteus         &                           & 34.78\% & 37.36\% & 40.18\% & 1.5493 \\ \bottomrule
        \end{tabular}}
    \caption{The prediction error for interference between concurrent kernels.}
    \label{table:performance_comparison}
\end{table}

Kernel overlap slowdown phenomenon has been overlooked by most existing simulation work.
Therefore, we mainly compare \sys with \bs, the only work we know that considers which simply uses a heuristic factor that varies only with GPU architecture and ML model to obtain the slowdown.

\noindent\textbf{Prediction accuracy.}
We evaluate slowdown prediction accuracy using a test dataset comprising approximately 5,000 kernels collected from various models. Table~\ref{table:performance_comparison} compares the interference prediction accuracy of \sys and \bs across different GPU types. 
\sys consistently achieves higher percentages of predictions within 5\%, 10\%, and 15\% error margins for the 3090 and H800 GPUs. Specifically, on the 3090 GPU, \sys attains 61.48\%, 73.62\%, and 76.52\% within 5\%, 10\%, and 15\% error respectively, significantly outperforming Proteus' 8.52\%, 13.62\%, and 19.28\%. On the H800 GPU, \sys achieves 43.18\%, 51.59\%, and 57.73\%, compared to Proteus' 34.78\%, 37.36\%, and 40.18\%. For the A800 GPU, \sys demonstrates competitive performance, achieving 61.03\% within 5\% error, closely matching Proteus' 59.04\%, while also surpassing in the 15\% error margin (75.01\% vs. 70.51\%). Additionally, \sys exhibits lower Root Mean Square Error (RMSE) values across all GPU types, with 1.0935, 0.7147, and 0.5626 for 3090, A800, and H800 respectively, compared to Proteus' 1.4953, 1.5170, and 1.5493. These results demonstrate that \sys provides more accurate slowdown predictions overall compared to the existing approach.

\begin{figure}[t]
    \centering 
    \includegraphics[width=0.95\linewidth]{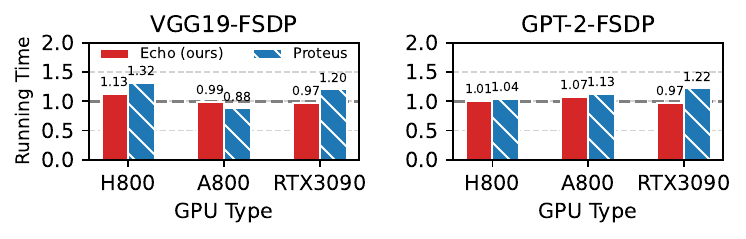} 
    \vspace{-4mm}
    \caption{
        The comparison of model-wise computation running time on different GPUs (H800, A800, RTX3090) in PyTorch FSDP mode for VGG19 and GPT-2 models. 
        Note that all the running times are normalized against the ground truth, represented by the gray line (y=1.0).
    }
    \label{fig:model_wise_overlap}
\end{figure}

\noindent\textbf{Model-wise evaluation.} We further assess the benefits of accurate slowdown prediction on model training simulations. Utilizing the advanced Fully Sharded Data Parallel (FSDP)~\cite{zhao2023pytorch} training mode from PyTorch, which incorporates computation-communication overlapping optimizations, we compare the overall computation time against ground truth and a baseline across different GPU architectures. As depicted in Figure~\ref{fig:model_wise_overlap}, \sys achieves an average slowdown prediction error rate of 4.67\%, significantly outperforming \bs, which has an error rate of 18.83\%. Notably, on the RTX3090 GPU, \sys maintains superior prediction accuracy, with its error rate being 8x lower than that of \bs.

\begin{figure}[t] 
    \centering 
    \includegraphics[width=\linewidth]{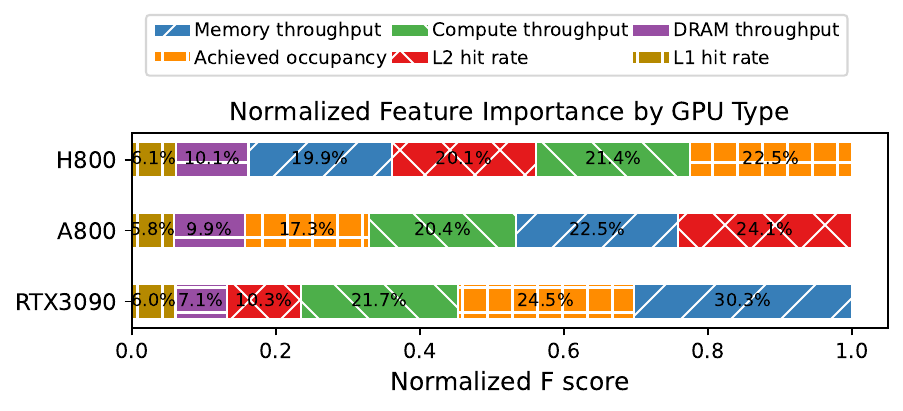} 
    \vspace{-6mm}
    \caption{The normalized feature importance of XGBoost across different GPU architectures, highlighting key computational resources.} 
    \label{fig:kernel_wise_compare} 
\end{figure}

\noindent\textbf{Feature score analysis.}
We further analyze the interpretability of \sys's XGBoost-based slowdown prediction model, analyzing feature importance across different GPU architectures. 
As illustrated in Figure~\ref{fig:kernel_wise_compare},
the features are ranked in ascending order based on their normalized importance.
For the RTX 3090, memory throughput emerges as the dominant feature, reflecting its bandwidth limitations due to the GDDR6 memory subsystem. The RTX 3090 has lower memory bandwidth compared to data-center GPUs, making it more prone to contention during overlapping kernels. This highlights the model's ability to identify memory throughput as the primary bottleneck.
On the A800, L2 cache hit rate and memory throughput are the most significant features, aligning with its optimized cache hierarchy. The A800's larger and more efficient L2 cache reduces DRAM latency, and the model's emphasis on cache-related metrics underscores the importance of effective cache utilization during kernel overlap scenarios. For the H800, the achieved SM occupancy and compute throughput score highest, while memory-related features are comparatively less significant. This reflects the Hopper architecture's focus on computational performance, including advanced SM partitioning for concurrent kernel execution. The lower relevance of memory-related features suggests that H800 is less constrained by memory bottlenecks, with performance under interference primarily driven by computational resource allocation and utilization.

\subsection{End-to-End Results} 
\label{subsec:overall_perf}

\begin{figure}[t]
    \centering 
    \includegraphics[width=1\linewidth]{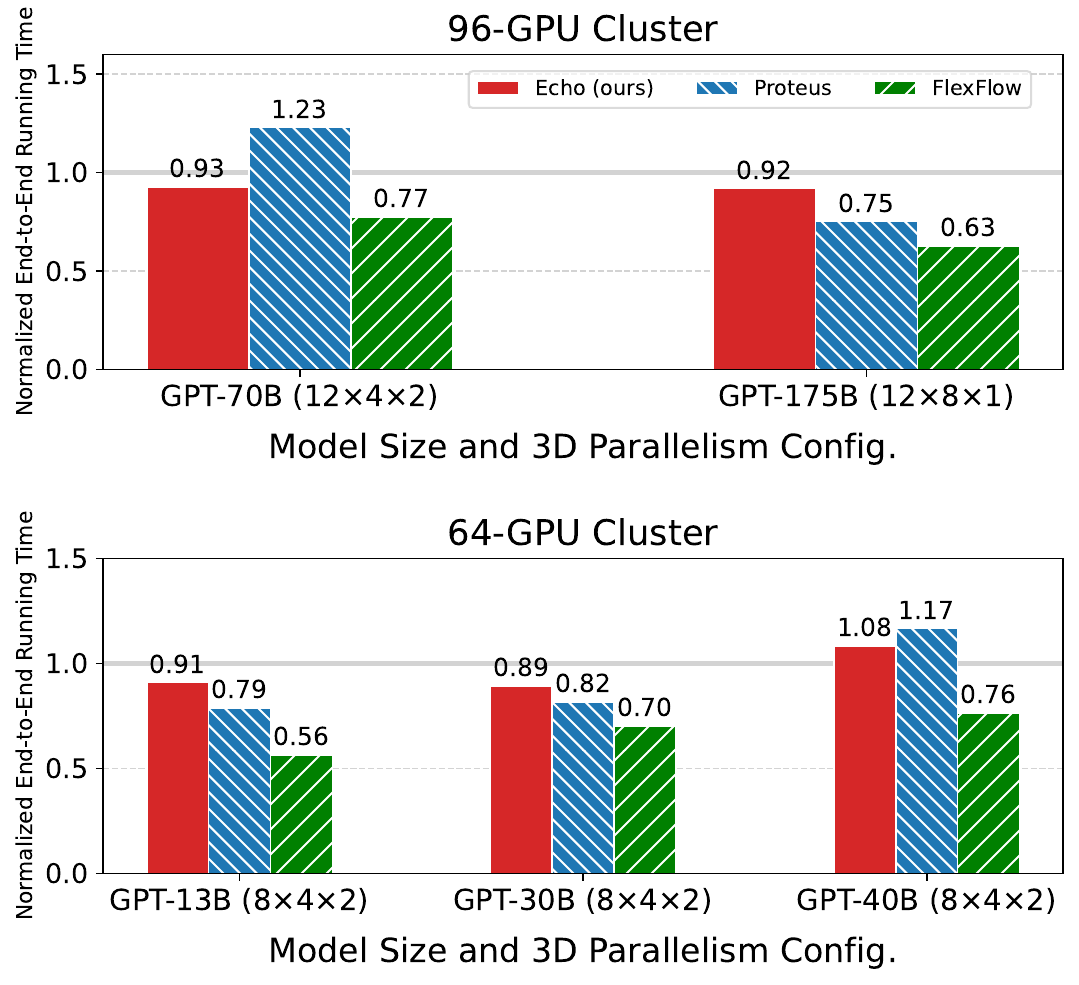}
    \caption{The end-to-end step time comparison for different GPT model sizes and 3D parallelism configurations (PP × TP × DP) on 64-GPU and 96-GPU H800 clusters. Note that all the data are normalized against the ground truth, represented by the gray line (y=1.0).} 
    \label{fig:end_to_end} 
\end{figure}

\noindent\textbf{Simulation accuracy.} 
We now evaluate the overall end-to-end simulation accuracy of \sys on H800 clusters of varying scales (with Megatron-LM). Figure~\ref{fig:end_to_end} compares \sys against baselines using GPT models of different sizes under diverse 3D parallelism configurations (PP × TP × DP).

Across all evaluated scenarios, \sys consistently aligns closely with real-world iteration times, substantially outperforming the baselines. For instance, on a 96-GPU H800 cluster, \sys's error for GPT-70B (12×4×2) and GPT-175B (12×8×1) is only 7\% and 8\%, respectively. In contrast, Proteus and FlexFlow exhibit higher deviations, with errors reaching up to 23\% and 25\% for Proteus, and 23\% and 37\% for FlexFlow on the same configurations. On a 64-GPU cluster, \sys achieves a 9\% error for GPT-13B (8×4×2), again surpassing both baselines.
Critically, \sys maintains an end-to-end error below 8.6\% even at larger scales. This high fidelity is attributed to \sys's accurate workload capture and communication modeling, including kernel overlap slowdown predictions. In contrast, the baselines rely on rigid assumptions and fail to incorporate framework-specific \op and communication optimizations, resulting in larger discrepancies in their predictions.

\noindent\textbf{Simulation time cost.} \sys incurs two primary types of time costs: (a) initialization time, which involves loading training workload trace files, initializing a series of variables, and enqueueing them into a waiting queue; and (b) execution time, which entails replaying the simulation by dequeuing \ops into timelines. Since the running time of computation \ops and workloads can be profiled and pre-traced and pre-traced for extended reuse, our evaluation focuses exclusively on simulation cost.
We assess two training modes—PyTorch's DDP and Megatron-LM's 3D parallelism—and report their corresponding average time costs in Table~\ref{table:simulation_cost}. In DDP mode, \sys creates a single copy of workloads for each DP group, resulting in stable and rapid average simulation times (averaging 10.3 seconds). In 3D parallelism mode, leveraging an effective white-box modeling approach for communication \ops, \sys achieves a 91.8× speedup compared to advanced simulators like SimAI on a 128-GPU setup (83s vs 7655s). Additionally, simulating an 8,192-GPU cluster requires only 1.38 hours with \sys.

\begin{table}[t]
\centering
\small
\resizebox{\linewidth}{!}{
\begin{tabular}{ccccccc}
\toprule
\multirow{2}{*}{\#GPUs} & \multicolumn{3}{c}{\textbf{VGG19}} & \multicolumn{3}{c}{\textbf{GPT 13B\textasciitilde 175B}} \\ \cmidrule(lr){2-4} \cmidrule(lr){5-7}
                        & \textbf{Init.} & \textbf{Execution} & \textbf{Total} & \textbf{Init.} & \textbf{Execution} & \textbf{Total} \\ \midrule
16                      & 0.3            & 9.5            & 9.8            & 0.3             & 4.1            & 4.4            \\
64                      & 0.4            & 10.3           & 10.6           & 2.3             & 40.1           & 42.4           \\
128                     & 0.3            & 10.0           & 10.3           & 4.3             & 79.0           & 83.4           \\
256                     & 0.4            & 10.3           & 10.7           & 8.4             & 159.2          & 167.6          \\
1024                    & 0.3            & 9.7            & 10.0           & 35.9            & 682.8          & 718.7          \\
4096                    & 0.4            & 10.2           & 10.6           & 149.9           & 3307.8         & 3457.7         \\
8192                    & 0.3            & 10.3           & 10.5           & 374.7           & 4602.2         & 4976.9         \\ \bottomrule
\end{tabular}}
\caption{\fyc{The simulation time costs of \sys in seconds. VGG19 is trained using the DDP mode, while GPT models (ranging from 13B to 175B) are trained using 3D parallelism.}}
\label{table:simulation_cost}
\end{table}

\section{Limitations and Discussions} 
\label{sec:discussion}

\sys has limitations and we wish to improve it in at least the following directions.

\noindent\textbf{Communication prediction.}
We plan to investigate learning based solutions like m3~\cite{li2024m3}, which efficiently predicts flow-level performance estimation while capturing network-centric factors such as transport protocols. 
We also wish to extend our modeling to explicitly consider network topology and the potential bandwidth contention, which is currently missing.

\noindent\textbf{Parallelism.}
\sys does not yet support expert parallelism for MoE, sequence/context parallelism, and memory-efficient ZeRO.
We plan to integrate them in subsequent releases.


\section{Related Work}
\label{sec:related}

We discuss related work other than those covered in \cref{sec:motivation}.

\noindent\textbf{GPU simulation.}
GPU computation simulators like SCALE-sim~\cite{samajdar2018scale} and Accel-Sim~\cite{khairy2020accel} focus on instruction-level modeling, while other approaches~\cite{hu2022dpro,lu2023distsim,luo2022srifty,zhu2020daydream} rely on profiling or tracing to capture accurate operator execution times. \sys similarly adopts a trace-based methodology, but crucially does so without requiring a full-scale deployment.

\noindent\textbf{Communication simulation.} 
Discrete event simulators like ns-3~\cite{riley2010ns} and OMNeT++~\cite{varga2019practical} are well-known to have high overheads due to their packet-level whole-stack simulation. Other than parallelization optimizations \cite{GCLL23,BZTW24}, recently learning based approach to predict packet-level or flow-level performance without simulating the whole stack~\cite{ZNKY21,li2024m3,YPCL22} has shown promising results.
As mentioned before, integrating them with \sys for training simulation is a promising direction for future work.

\section{Conclusion}
We presented \sys, a high-fidelity simulator for large-scale distributed ML training that bridges critical gaps in existing approaches. 
\sys simulates without full-scale deployment with ex-situ workload tracing, estimates CC performance via white-box modeling, and accounts for the slowdown caused by kernels overlapping.
Our evaluation shows that \sys not only achieves up to 3x lower simulation error than state-of-the-art baselines but also is highly efficient and practical.
We plan to open source \sys to the community.  

\section*{Acknowledgments}
We thank Hongming Huang, Cheng Luo, Jiamin Li, Ran Shu and Lei Qu for their help in the early stage of this project; Kin Hang Sew and Boying Chen for their help with the implementation and experiments.
We also thank Jiaqi Gao and Jiangfei Duan for very help discussions.

\clearpage
\bibliographystyle{plain}
\bibliography{ref}
\clearpage

\appendix

\section{Additional Observations}
\label{sec:additional_obs}
We present additional observations on distributed model training in 256-GPU A100 clusters.
These cluster consist of up to 256 GPUs (32 servers), with each server equipped with 8 Ampere A100 40GB Tensor Core GPUs. The intra-server connections utilize NVLink, providing 600 GBps of bandwidth, while the inter-server network offers each GPU a dedicated 200 Gb/s InfiniBand connection.

\subsection{Training Efficiency Gap}
\begin{figure}[t] 

    \begin{minipage}{0.4\textwidth} 
        \centering 
        \includegraphics[width=0.9\linewidth] {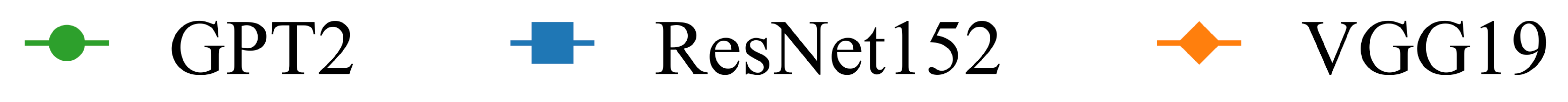} 
    \end{minipage} %
    \hfill
    \begin{subfigure}[t]{0.22\textwidth} 
        \centering 
        \includegraphics[width=0.8\linewidth] {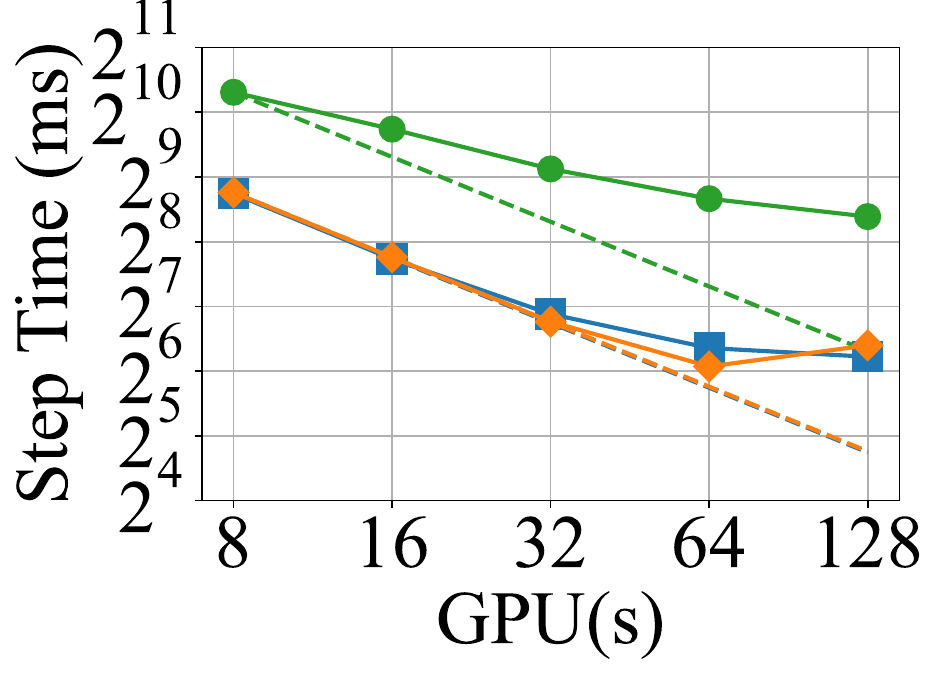} 
        \caption{} 
        \label{fig:motivation_cluster_scale} 
    \end{subfigure} 
    \begin{subfigure}[t]{0.24\textwidth} 
        \centering 
        \includegraphics[width=0.8\linewidth] {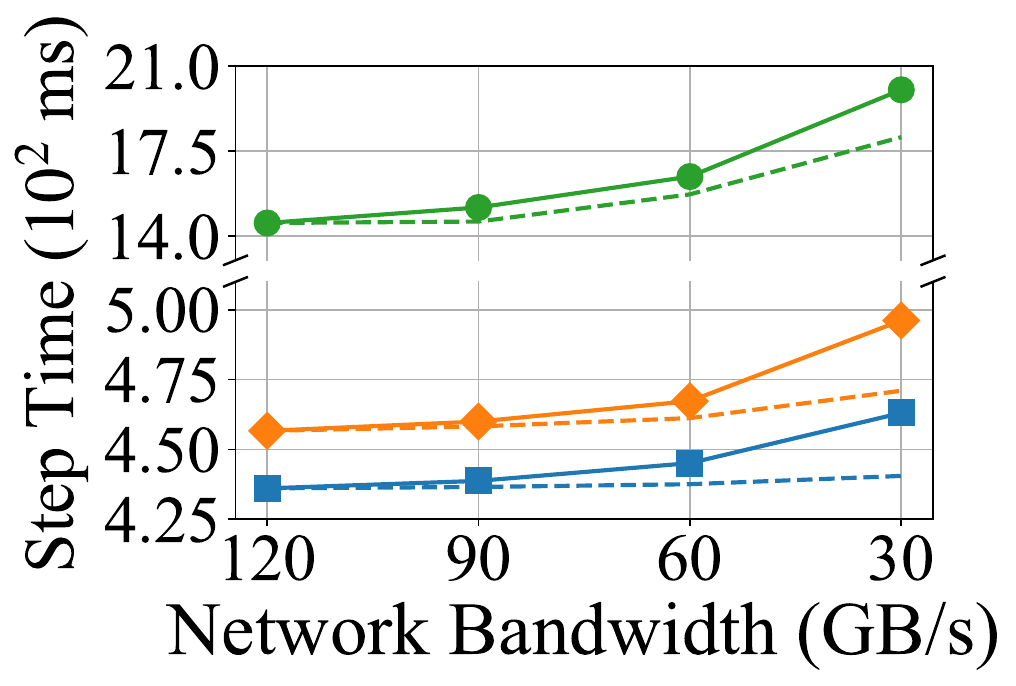} 
        \caption{} 
        \label{fig:motivation_network} 
    \end{subfigure} 
    \vspace{-3mm} 
    \footnotesize
    \caption{Step time change of three DNN models profiled on A100 cluster. The dashed lines denote the theoretical performance. For network bandwidth, we linearly scale all the communication time and compute the theoretical step time.} 
    \label{fig:perf_gap} 
    \vspace{-3mm}
\end{figure}

Training time cannot be easily predict. Numerous approaches have been proposed to accelerate training, with practitioners aiming for linear speedup as more devices are added. However, frequent inter-device communication makes achieving this efficiency challenging, as highlighted by extensive research and our own empirical observations. Figure~\ref{fig:perf_gap} illustrates the performance of three classic DNN models across different cluster sizes and network bandwidths. As cluster size increases, the step time deviates from the expected speedup, with a maximum degradation of 33.1\% (Figure~\ref{fig:motivation_cluster_scale}). Furthermore, scaling communication time linearly with network bandwidth fails to predict the actual step time (Figure~\ref{fig:motivation_network}).

\subsection{Communication Analysis}
\begin{figure}[t] 
    \begin{subfigure}[t]{0.23\textwidth} 
        \centering 
        \includegraphics[width=\linewidth] {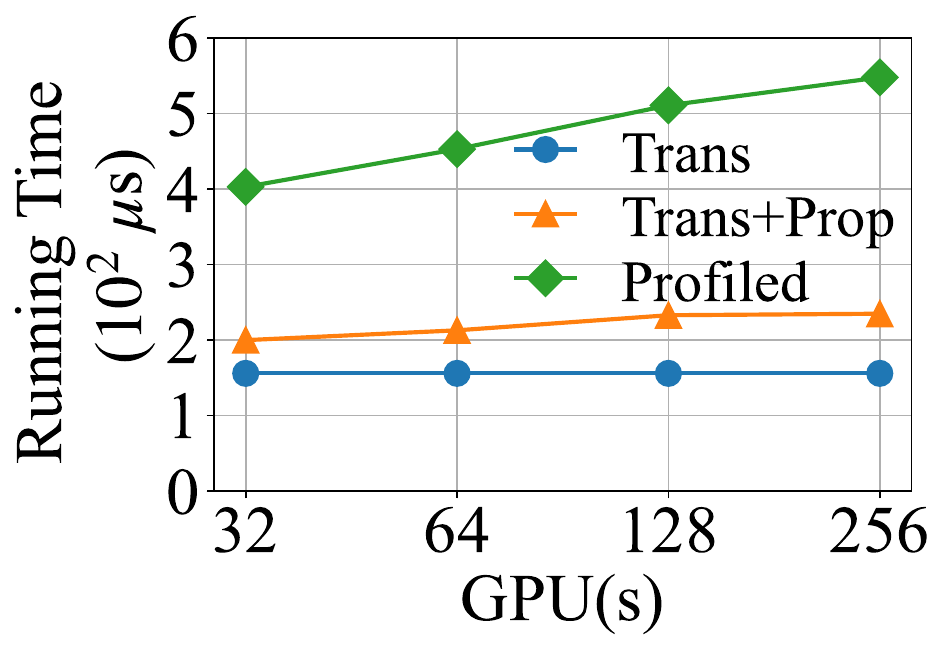}
        \caption{Gap with profiled running time.} \label{fig:transmission_nccl_gap} 
    \end{subfigure}%
    \hfill 
    \begin{subfigure}[t]{0.23\textwidth} 
        \centering 
        \includegraphics[width=\linewidth] {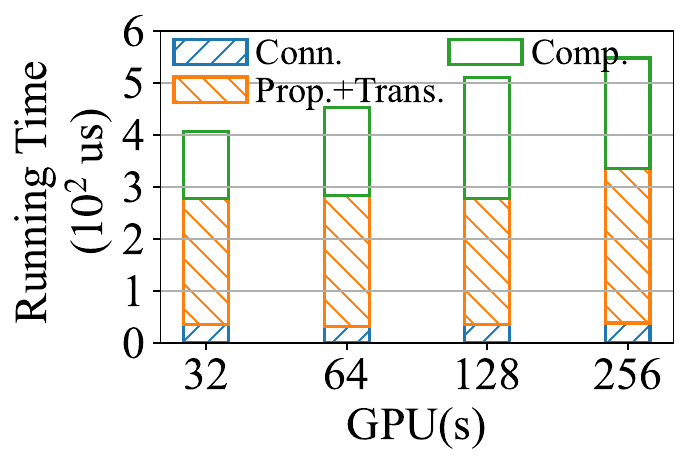} 
        \caption{Breakdown of running time.} 
        \label{fig:nccl_breakdown} 
    \end{subfigure} 
    \vspace{-1mm} 
    \caption{Running time of NCCL All-Reduce with varying GPU counts, profiled on A100 clusters. We use NCCL-test to launch all-reduce kernels and profile the end-to-end running time. } 
    \vspace{-1mm}
    \label{fig:nccl_perf_gap} 
\end{figure}

\noindent\textbf{Theoretical model analysis.}
We empirically show the running time gap between the theoretical transmission delay and the true running time of the all-reduce kernel. We use NCCL-test to launch all-reduce kernels and profile the end-to-end running time. Figure~\ref{fig:transmission_nccl_gap} depicts the performance in different cluster size. The value of $tensor\_size/bandwidth$ is a constant while the running time of all-reduce gradually grows with the cluster size. In a cluster with 256 GPUs, the performance gap is up to 251\%. Then, we consider another possible solution, Trans + Prop in Figure~\ref{fig:transmission_nccl_gap}. It incorporates theoretical propagation delay ($distance/propagation\_speed$) because servers propagate reduced tensor to others in the all-reduce kernel. However, it cannot simulate correctly either.

\noindent\textbf{NCCL breakdown analysis.}
We thus perform a breakdown analysis of the all-reduce running time in Figure~\ref{fig:nccl_breakdown}. A complete all-reduce kernel involves (a) initiating the connection, (b) transmitting and propagating the message among servers and (c) averaging the results. Empirically, we find that the running time of all-reduce kernel is indeed a combination of the transfer time, data reduction time and overhead of connection setup. Here, the transfer time is measured to be the time elapsed between the start of the transmission on the sender and the end of receiving the tensor on the receiver. We notice that an average cost of 352.5\,$\mu$s is spent on connection setup. Both propagation and computation time for reducing the tensor are non-trivial and gradually increase as the cluster scales up. Apart from the significant contribution of reduce computation, the profiled transfer time is larger than the sum of the theoretical transmission delay and propagation delay.

\section{Model Configurations}
\label{sec:model_configurations}
The GPT model specifications are summarized in Table~\ref{table:gpt_config}.

\begin{figure*}[t] 
    \centering 
    \includegraphics[width=1\linewidth]{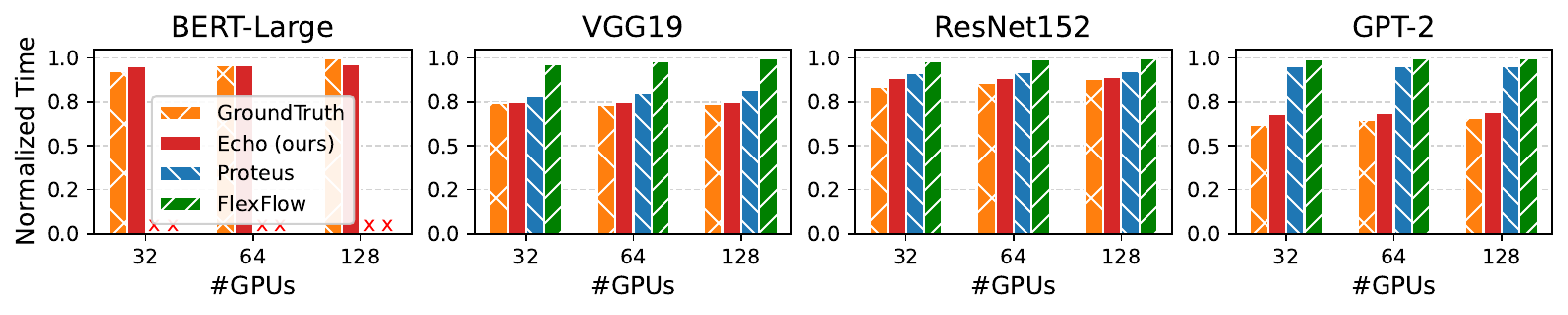} 
    \vspace{-7mm}
    \caption{Comparison of prediction accuracy between Baseline and \sys when the cluster size increases.} 
    \label{fig:ddp128_compare}
    \vspace{-3mm}
\end{figure*}

\section{CDFs of Slowdown Predictions}
\begin{figure}[t] 
    \centering 
    \includegraphics[width=0.8\linewidth]{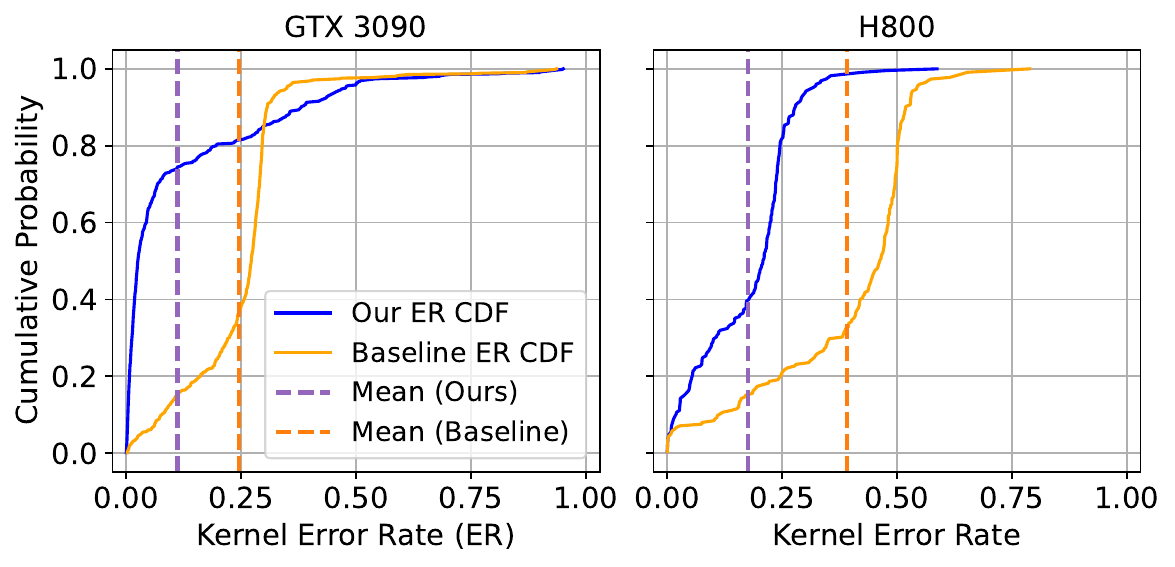} 
    \vspace{-2mm}
    \caption{CDFs of kernel slowdown prediction error rate predictions for \sys and the baseline (Proteus) on RTX 3090 and H800 cluster.} 
    \label{fig:cdf_slowdown} 
\end{figure}

Figure~\ref{fig:cdf_slowdown} compares the cumulative distribution functions (CDFs) of kernel slowdown prediction error rate predictions made by \sys and the baseline method, Proteus. The comparison is shown for two hardware setups: an RTX 3090 GPU and an H800 cluster. \sys demonstrates a significantly lower kernel error rate across both platforms, with a sharper CDF slope and lower mean error rate compared to the baseline.

\begin{table}[b]
\small
\hfill
\begin{minipage}{1\linewidth} 
\centering
\resizebox{\linewidth}{!}{
\begin{tabular}{l l c c c c}
\toprule
Model  & \#Params   & \#Heads & \#Layers & \#Hidden Size & \#Sequence Length \\ \midrule
GPT    & 13B        & 32      & 40       & 5120          & 2048               \\ 
GPT    & 30B        & 48      & 40       & 7680          & 2048               \\ 
GPT    & 40B        & 72      & 40       & 9216          & 2048               \\ 
GPT    & 70B        & 64      & 80       & 8192          & 2048               \\ 
GPT    & 175B       & 96      & 96       & 12288         & 2048               \\ \bottomrule
\end{tabular}}
\caption{Model configurations for GPT series.}
\label{table:gpt_config}
\end{minipage}
\end{table}

\section{End-to-End Simulation}
We evaluate the accuracy of \sys in different training configurations under PyTorch's DDP mode. We adjust the cluster size and network bandwidth. Figure~\ref{fig:ddp128_compare} shows the performance of the baseline and \sys compared with the profiled ones. The average prediction accuracy is 97.25\%. For language models, the error rate is reduced by at least 2.02x. We find that the Baseline solution performs better in CNN models.
We also evaluate \sys by replacing the InfiniBand network with a 100GB/s PCIe interconnection so that it takes a long time to transmit model gradients (Table~\ref{tab:prediction_accuracy_network}). 
The average accuracy is 96.17\%. Degradation in the accuracy mainly stems from the error in predicting the slowdown factor.

\begin{table}[t]
\centering
\small
\resizebox{\linewidth}{!}{
                \begin{tabular}{@{}lllll@{}}
                \toprule
                Model                          & Network (/s) & Profiled (ms) & Prediction (ms) & Error (\%) \\ \midrule
                \multirow{2}{*}{BERT\_Large}   & 100Gb   & 454      & 470.8      & 3.70      \\
                                               & 1.6Tb   & 355.8    & 376.6      & 5.84      \\ \midrule
                \multirow{2}{*}{GPT2}          & 100Gb   & 199.5    & 206.3      & 3.40      \\
                                               & 1.6Tb   & 167.4    & 176.9      & 5.67       \\ \midrule
                \multirow{2}{*}{VGG19}         & 100Gb   & 529.9    & 521.4      & 1.60       \\
                                               & 1.6Tb   & 485.6    & 489.2      & 0.86       \\ \midrule
                \multirow{2}{*}{ResNet152}     & 100Gb   & 351.1    & 374.3      & 6.61       \\
                                               & 1.6Tb   & 335.7    & 354.7      & 5.65       \\ \bottomrule
                \end{tabular}
}
\caption{\fyc{Prediction error rate of \sys in the cluster with 128 A100 GPUs. We change the network interconnection from 1.6Tb/s InfiniBand to 100 Gb/s PCIe.}}
\label{tab:prediction_accuracy_network}
\end{table}

\end{document}